\documentclass{article}

\usepackage{microtype}
\usepackage{graphicx}
\usepackage{subcaption}
\usepackage{booktabs} % for professional tables

\usepackage{hyperref}

\usepackage[accepted]{icml2024}

\usepackage{amsmath}
\usepackage{amssymb}
\usepackage{mathtools}
\usepackage{amsthm}

\usepackage[capitalize,noabbrev]{cleveref}

\theoremstyle{plain}

\theoremstyle{definition}

\theoremstyle{remark}

\usepackage{caption}
\usepackage{algorithm}
\usepackage{algorithmic}
\usepackage{graphicx}
\usepackage{multirow}
\usepackage{tabularx}
\usepackage{wrapfig}
\usepackage[nottoc]{tocbibind}
\usepackage{minitoc}
\usepackage{bbm}

\newcommand{\mytitle}{Diffusion Model-Augmented Behavioral Cloning}

\newcommand{\sun}[1]{{\color{blue}{\small\bf\sf [Sun: #1]}}}

\newcommand{\Skip}[1]{}

\newcommand{\method}{DBC}
\newcommand{\methodFull}{Diffusion Model-Augmented Behavioral Cloning}
\newcommand{\maze}{\textsc{Maze}}
\newcommand{\fetchpick}{\textsc{FetchPick}}

\newcommand{\handrotate}{\textsc{HandRotate}}
\newcommand{\walker}{\textsc{Walker}}
\newcommand{\antreach}{\textsc{AntReach}}
\newcommand{\cheetah}{\textsc{Cheetah}}
\newcommand{\carracing}{\textsc{CarRacing}}

\newcommand{\ie}{\textit{i}.\textit{e}.,\ }
\newcommand{\eg}{\textit{e}.\textit{g}.,\ }

\newcommand{\myfig}[1]{Figure \ref{#1}}
\newcommand{\mytable}[1]{Table \ref{#1}}
\newcommand{\myeq}[1]{Eq. \ref{#1}}
\newcommand{\mysecref}[1]{Section \ref{#1}}
\newcommand{\myalgo}[1]{Algorithm \ref{#1}}

\newcommand{\dotieconcat}[2]{% auxiliary macro, don't use it directly
  \text{\raisebox{.8ex}{$\smallfrown$}}%
}

\makeatletter
\newcommand\dslfontsize{\@setfontsize\dslfontsize\@viipt\@viiipt}
\renewcommand\scriptsize{\@setfontsize\subfigcap{7}{8}}%
\makeatother

\newcommand{\myparagraph}[1]{\noindent \textbf{#1.}}

\usepackage{color}
\usepackage{xcolor}
\definecolor{codegray}{rgb}{0.5,0.5,0.5}
\newcommand{\argmaxaShort}{$\arg\max_{a \in \mathcal{A}}\ p(s, a)$}

\usepackage{amsmath,amsfonts,bm}

\def\eqref#1{equation~\ref{#1}}
\def\1{\bm{1}}

\DeclareMathAlphabet{\mathsfit}{\encodingdefault}{\sfdefault}{m}{sl}
\SetMathAlphabet{\mathsfit}{bold}{\encodingdefault}{\sfdefault}{bx}{n}

\usepackage[textsize=tiny]{todonotes}

\icmltitlerunning{\mytitle}

\begin{document}
\doparttoc % Tell to minitoc to generate a toc for the parts
\faketableofcontents % Run a fake tableofcontents command for the partocs

\twocolumn[
\icmltitle{\mytitle}



\icmlsetsymbol{equal}{*}

\begin{icmlauthorlist}
\icmlauthor{Shang-Fu Chen}{equal,ntu}
\icmlauthor{Hsiang-Chun Wang}{equal,ntu}
\icmlauthor{Ming-Hao Hsu}{ntu}
\icmlauthor{Chun-Mao Lai}{ntu}
\icmlauthor{Shao-Hua Sun}{ntu}
\end{icmlauthorlist}

\icmlaffiliation{ntu}{National Taiwan University, Taipei, Taiwan}

\icmlcorrespondingauthor{Shang-Fu Chen}{f07942144@ntu.edu.tw}
\icmlcorrespondingauthor{Shao-Hua Sun}{shaohuas@ntu.edu.tw}

\icmlkeywords{Machine Learning, ICML}

\vskip 0.3in
]



\printAffiliationsAndNotice{\icmlEqualContribution} % otherwise use the standard text.
\begin{abstract}
Imitation learning addresses the challenge of learning by observing an expert’s demonstrations without access to reward signals from environments. 
Most existing imitation learning methods that do not require interacting with environments  
either model the expert distribution 
as the conditional probability
$p(a|s)$ (\eg behavioral cloning, BC)
or the joint probability $p(s,a)$.
Despite the simplicity of modeling the conditional probability with BC, it usually struggles with generalization.
While modeling the joint probability can improve generalization performance, 
the inference procedure is often time-consuming, and the model can suffer from manifold overfitting.
This work proposes an imitation learning framework that benefits from modeling both the conditional and joint probability of the expert distribution.
Our proposed \methodFull{} (\method{}) employs a diffusion model trained to model expert behaviors and learns a policy to optimize both the BC loss (conditional) and our proposed diffusion model loss (joint).
\method{} outperforms baselines in various continuous control tasks in navigation, robot arm manipulation, dexterous manipulation, and locomotion. 
We design additional experiments to verify the limitations of modeling either the conditional probability or the joint probability of the expert distribution, as well as compare different generative models. 
Ablation studies justify the effectiveness of our design choices. % \blfootnote{Project page: \url{https://nturobotlearninglab.github.io/dbc}}
\end{abstract}

\section{Introduction}
\label{sec:intro}

Recently, the success of deep reinforcement learning (DRL)~\citep{mnih2015human, LillicrapHPHETS15, arulkumaran2017deep}
has inspired the research community to develop DRL frameworks to control robots, aiming to automate the process of designing sensing, planning, and control algorithms by letting the robot learn in an end-to-end fashion. 
Yet, acquiring complex skills through trial and error can still lead to undesired behaviors even with sophisticated reward design~\citep{christiano2017deep, leike2018scalable, lee2019composing}. 
Moreover, the exploring process could damage expensive robotic platforms or even be dangerous to humans~\citep{garcia2015comprehensive, levine2020offline}.

To overcome this issue, imitation learning (\ie learning from demonstration)~\citep{schaal1997learning, osa2018algorithmic} has received growing attention, whose aim is to learn a policy from 
expert demonstrations, which are often more accessible than appropriate reward functions for reinforcement learning. 
Among various imitation learning directions, adversarial imitation learning~\citep{ho2016generative, zolna2021task, kostrikov2018discriminatoractorcritic} and inverse reinforcement learning~\citep{ng2000algorithms, abbeel2004apprenticeship} have achieved encouraging results in a variety of domains. Yet, these methods require interacting with environments, which can still be expensive or even dangerous.

On the other hand, behavioral cloning (BC)~\citep{pomerleau1989alvinn, bain1995framework} does not require interacting with environments.
BC formulates imitation learning as a supervised learning problem --- given an expert demonstration dataset, an agent policy takes states sampled from the dataset as input and learns to replicate the corresponding expert actions. 
One can view a BC policy as a discriminative model $p(a|s)$ that models the \textit{conditional probability} of actions $a$ given a state $s$.
Due to its simplicity and training stability, BC has been widely adopted for various applications.
However, BC struggles at generalizing to states unobserved during training~\citep{nguyen2023reliable}.

To alleviate the generalization issue, we propose to augment BC by modeling the \textit{joint probability} $p(s, a)$ of expert state-action pairs with a generative model (\eg diffusion models). 
This approach is motivated by~\citet{bishop2006pattern} and ~\citet{fisch2013knowledge}, who illustrate that modeling joint probability allows for better generalizing to data points unobserved during training.
However, with a learned joint probability model $p(s, a)$, retrieving a desired action $a$ requires actions sampling and optimization, \ie $\underset{a \in \mathcal{A}}{\arg\max}\ p(s, a)$, which can be extremely inefficient with a large action space.
Moreover, modeling joint probabilities can suffer from manifold overfitting~\citep{wu2021bridging, loaiza2022diagnosing} when observed high-dimensional data lies on a low-dimensional manifold  (\eg state-action pairs collected from a script expert policies).

This work proposes an imitation learning framework that combines both the efficiency and stability of modeling the \textit{conditional probability}  and the generalization ability of modeling the \textit{joint probability}. Specifically, we propose to model the expert state-action pairs using a state-of-the-art generative model, a diffusion model, which learns to estimate how likely a state-action pair is sampled from the expert dataset. Then, we train a policy to optimize both the BC objective and the learning signals the trained diffusion model produces. Therefore, our proposed framework not only can efficiently predict actions given states via capturing the \textit{conditional probability} $p(a|s)$ but also enjoys the generalization ability induced by modeling the \textit{joint probability} $p(s, a)$ and utilizing it to guide policy learning.
 
We evaluate our proposed framework and baselines in various continuous control domains, including navigation, robot arm manipulation, and locomotion. 
The experimental results show that the proposed framework outperforms all the baselines or achieves competitive performance on all tasks. 
Extensive ablation studies compare our proposed method to its variants,
justifying our design choices, such as different generative models, and investigating the effect of hyperparameters.
\section{Related Work}
\label{sec:related_work}
Imitation learning aims to learn by observing expert demonstrations without access to rewards from environments.
It has various applications such as robotics~\citep{schaal1997learning, sun2018neural, zhao2023learning}, autonomous driving~\citep{Codevilla2020drive}, and game AI~\citep{harmer2018game}. 

\myparagraph{Behavioral Cloning (BC)} 
BC and its extensions~\citep{pomerleau1989alvinn, torabi2018behavioral, shafiullah2022behavior, zhao2023learning} 
formulates imitating an expert as a supervised learning problem. 
Due to its simplicity and effectiveness, 
it has been widely adopted in various domains.
Yet, it often struggles at generalizing to states unobserved from the expert demonstrations. 
To alleviate the above problem,~\citet{ross2011reduction} propose the DAgger algorithm that gradually accumulates additional expert demonstrations to mitigate the deviation from the expert, which relies on the availability of querying an expert;
Implicit BC (IBC) \citep{florence2022implicit} demonstrates better generalization than BC by using an energy-based model for state-action pairs. However, it requires time-consuming action sampling and optimization during inference, which may not scale well to high-dimensional action spaces.
In this work, we improve the generalization ability of policies by augmenting BC with a diffusion model that learns to capture the joint probability of expert
state-action pairs.

\myparagraph{Adversarial Imitation Learning (AIL)}
AIL methods aim to match the state-action distributions of an agent and an expert via adversarial training. 
Generative adversarial imitation learning (GAIL)~\citep{ho2016generative} 
and its extensions~\citep{torabi2018generative, kostrikov2018discriminatoractorcritic, zolna2021task, jena2021augmenting, lai2024diffusion}
resemble the idea of generative adversarial networks~\citep{goodfellow2014generative}, 
which trains a generator policy to imitate expert behaviors and a discriminator 
to distinguish between the expert and the learner's state-action pair distributions. 
While modeling state-action distributions often leads to satisfactory performance,
adversarial learning can be unstable and inefficient~\citep{Chen2020On}. Moreover, even though scholars like ~\citet{jena2021augmenting} propose to improve the efficiency of GAIL with the BC loss, they still require online interaction with environments, which can be costly or even dangerous. In contrast, our work does not require interacting with environments.

\myparagraph{Inverse Reinforcement Learning (IRL)} 
IRL methods~\citep{ng2000algorithms, abbeel2004apprenticeship, ziebart2008maximum, fu2018learning, goalprox, swamy2023inverse} are designed to infer the reward function that underlies the expert demonstrations
and then learn a policy using the inferred reward function.
This allows for learning tasks whose reward functions are difficult to specify manually.
However, due to its double-loop learning procedure, IRL methods are typically computationally expensive and time-consuming.
Additionally, obtaining accurate estimates of the expert's reward function can be difficult, especially when the expert's behavior is non-deterministic or when the expert's demonstrations are sub-optimal. 

\myparagraph{Diffusion Policies} 
Recently, \citet{pearce2022imitating, chi2023diffusionpolicy, reuss2023goal}
propose to represent and learn an imitation learning policy using a conditional diffusion model, 
which produces a predicted action conditioning on a state and a sampled noise vector.
These methods achieve encouraging results in modeling stochastic and multimodal behaviors from human experts or play data.
In contrast, instead of representing a policy using a diffusion model, our work employs a diffusion model trained on expert demonstrations to guide a policy as a learning objective.
\section{Preliminaries}
\label{sec:prelim}

\subsection{Imitation Learning}
\label{sec:prelim_il}

In contrast to reinforcement learning, whose goal is to learn a policy $\pi$ based on rewards received while interacting with the environment,  
imitation learning methods aim to learn the policy from an expert demonstration dataset containing $M$ trajectories, $D=\{\tau_{1},...,\tau_{M}\}$, where $\tau_i$ represents a sequence of $n_i$ state-action pairs $\{s^i_1, a^i_1, ..., s^i_{n_i}, a^i_{n_i}\}$.

\subsubsection{Modeling Conditional Probability $p(a|s)$}
\label{sec:prelim_bc}

To learn a policy $\pi$,
behavioral cloning (BC) directly estimates the expert policy $\pi^{E}$
with maximum likelihood estimation (MLE).
Given a state-action pair $(s, a)$ sampled from the dataset $D$,
BC optimizes
$\mathop{max}\limits_{\theta} \sum\limits_{(s,a)\in D} \log (\pi_{\theta}(a|s))$,
where $\theta$ denotes the parameters of the policy $\pi$.
One can view a BC policy as a discriminative model $p(a|s)$, capturing the \textit{conditional probability} of an action $a$ given a state $s$.
On the other hand, Implicit BC~\citep{florence2022implicit, ganapathi2022implicit} propose to model the conditional probability with InfoNCE-style~\citep{oord2018representation} optimization.
Despite their success in various applications,
BC-based methods tend to overfit and struggle at generalizing to 
states unseen during training~\citep{ross2011reduction, codevilla2019exploring, wang2022high}.

\subsubsection{Modeling Joint Probability $p(s, a)$}
\label{sec:prelim_state_action}
In order to model the \textit{joint probability} $p(s, a)$ of the expert dataset for improved generalization performance~\citep{bishop2006pattern, fisch2013knowledge}, one can employ explicit generative models, such as 
energy-based models~\citep{du2019implicit, song2021train}, 
variational autoencoders~\citep{kingma2013auto}, 
and flow-based models~\citep{rezende2015variational, dinh2017density}.
However, these methods can be extremely inefficient in retrieving actions with a large action space during inference since sampling and optimizing actions (\ie \argmaxaShort{}) are required.
Moreover, they are known to struggle with modeling 
observed high-dimensional data that
lies on a low-dimensional manifold 
(\ie manifold overfitting)~\citep{wu2021bridging, loaiza2022diagnosing}.
As a result, these methods often perform poorly when learning from demonstrations 
produced by script policies or PID controllers, as discussed in~\mysecref{sec:conditional_vs_joint}.

We aim to develop an imitation learning framework that enjoys the advantages of modeling the \textit{conditional probability} $p(a|s)$ and the \textit{joint probability} $p(s, a)$. Specifically, we propose to model the \textit{joint probability} of expert state-action pairs using an explicit generative model $\phi$, which learns to produce an estimate indicating how likely a state-action pair is sampled from the expert dataset. Then, we train a policy to model the \textit{conditional probability} $p(a|s)$ by optimizing the BC objective and the estimate produced by the learned generative model $\phi$. Hence, our method can efficiently predict actions given states, generalize better to unseen states, and suffer less from manifold overfitting.

\subsection{Diffusion Models}
\label{sec:prelim_diff}
% % \vspace{-1.5cm}
% \begin{wrapfigure}[14]{r}{0.47\textwidth}
%     % \vspace{-2cm}
%     \begin{center}
%     \includegraphics[width=\linewidth]{figure/diffusion_model_fig.pdf}
%     \end{center}
%     \vspace{-0.18cm}    
% \caption[]{\textbf{Denoising Diffusion Probabilistic Model (DDPM).} 
% Latent variables $x_1, ..., x_N$ are produced from the data point $x_0$ via the forward diffusion process, \ie gradually adding noises to the latent variables. The diffusion model $\phi$ learns to reverse the diffusion process by denoising the noisy data to reconstruct the original data point $x_0$.
% }
% % \task{caption}
% \label{fig:dm}
% \end{wrapfigure}

\begin{figure}[t]
    \centering
    \includegraphics[width=0.47\textwidth]{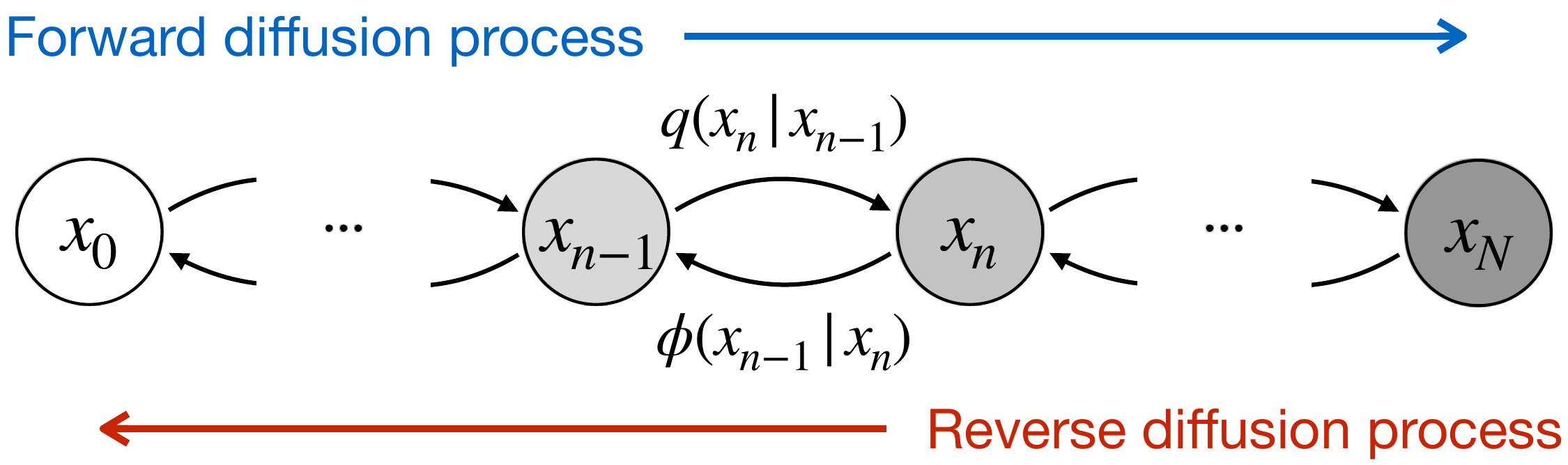}
    \vspace{-0.4cm}
    \caption[]{\textbf{Denoising Diffusion Probabilistic Model (DDPM).} 
    Latent variables $x_1, ..., x_N$ are produced from the data point $x_0$ via the forward diffusion process, \ie gradually adding noises to the latent variables. The diffusion model $\phi$ learns to reverse the diffusion process by denoising the noisy data to reconstruct the original data point $x_0$.
    }
    \label{fig:dm}
\end{figure}
As described in the previous sections, 
this work aims to combine the advantages of modeling the \textit{conditional probability} $p(a|s)$ and the \textit{joint probability} $p(s, a)$.
Hence, we leverage diffusion models
to model the \textit{joint probability} of expert state-action pairs.
The diffusion model is a recently developed class of generative models and has achieved state-of-the-art performance
on various tasks~\citep{dickstei2015thermodynamics, nichol2021improved, dhariwal2021diffusion, ko2023learning, poole2023dreamfusion}.

In this work, 
we utilize Denoising Diffusion Probabilistic Models (DDPMs)~\citep{ho2020ddpm} to model expert state-action pairs. Specifically, DDPM models gradually add noise to 
data samples (\ie concatenated state-action pairs)
until they become isotropic Gaussian (\textit{forward diffusion process}),
and then learn to denoise each step and restore the original data samples (\textit{reverse diffusion process}),
as illustrated in~\myfig{fig:dm}.
In other words, DDPM learns to recognize a data distribution by learning to denoise noisy sampled data.
More discussion on the relationship between diffusion models and the data distribution can be found in~\mysecref{sec:app_diffusion_model}.
\begin{figure*}[t]
\centering
    \includegraphics[width=\textwidth]{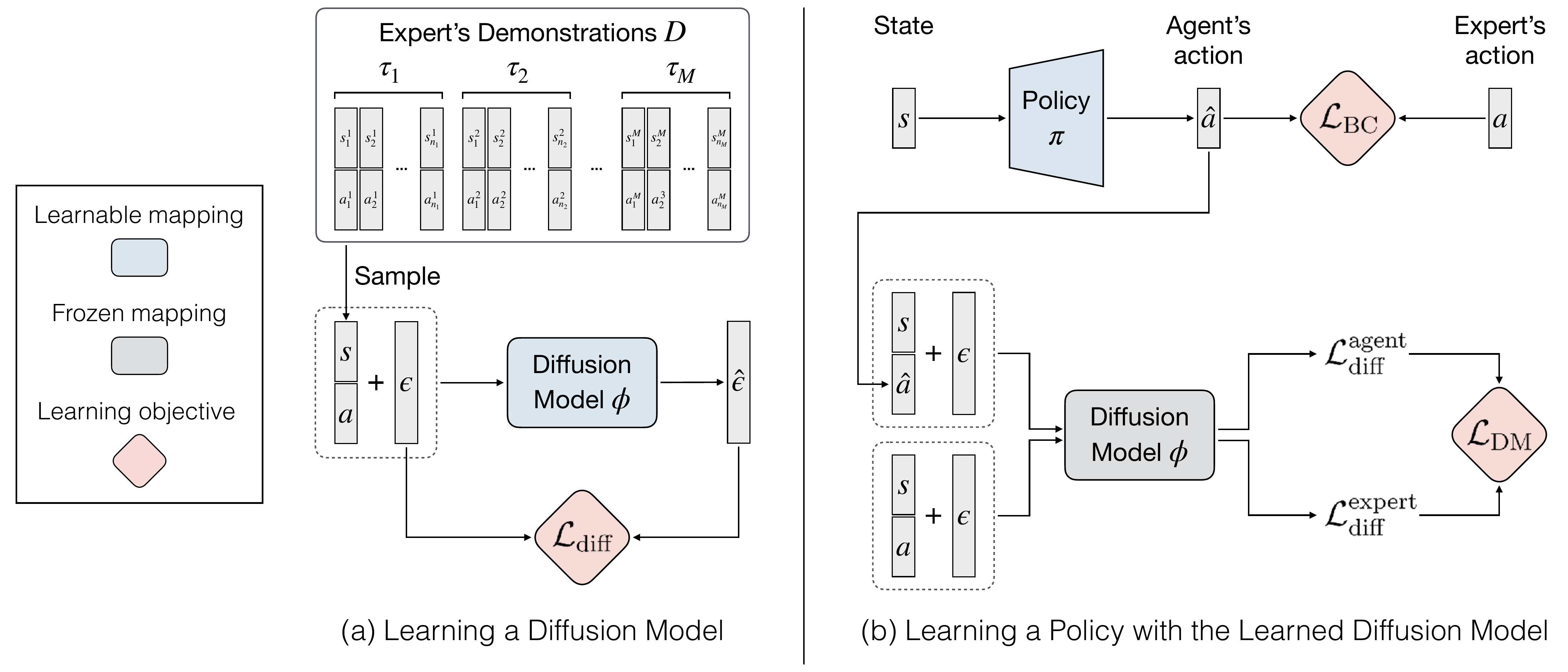} % {figure/model.png}

    \vspace{-0.4cm}
    \caption[]{
        \small 
        \textbf{\methodFull{} (\method{} ).} Our proposed framework augments behavioral cloning (BC) by employing a diffusion model.
        % Specifically, it first learns a diffusion model $\phi$ from the expert's demonstrations by optimizing $\mathcal{L}_\text{diff}$. Then, it learns a policy $\pi$ by optimizing the BC objective $\mathcal{L}_\text{BC}$ together with our proposed diffusion model objective $\mathcal{L}_\text{DM}$.
        (a) \textbf{Learning a Diffusion Model}: the diffusion model $\phi$ learns to model the distribution of concatenated state-action pairs sampled from the demonstration dataset $D$. It learns to reverse the diffusion process (\ie denoise) by optimizing $\mathcal{L}_\text{diff}$ in~\myeq{eq:diff_loss}.
        (b) \textbf{Learning a Policy with the Learned Diffusion Model}: we propose a diffusion model objective $\mathcal{L}_{\text{DM}}$ for policy learning and jointly optimize it with the BC objective $\mathcal{L}_{\text{BC}}$. Specifically, $\mathcal{L}_{\text{DM}}$ is computed based on processing a sampled state-action pair $(s, a)$ and a state-action pair $(s, \hat{a})$ with the action $\hat{a}$ predicted by the policy $\pi$ with $\mathcal{L}_\text{diff}$.
        
        \label{fig:model}
    }
\end{figure*}
\section{Approach}
\label{sec:approach}
Our goal is to design an imitation learning framework that enjoys both the advantages of modeling the \textit{conditional probability} and the \textit{joint probability} of expert behaviors.
To this end, we first adopt behavioral cloning (BC) for modeling the \textit{conditional probability} from expert state-action pairs, as described in~\mysecref{sec:bc_learning}.
To capture the \textit{joint probability} of expert state-action pairs, we employ a diffusion model that learns to produce an estimate indicating how likely a state-action pair is sampled from the expert state-action pair distribution, as presented in~\mysecref{sec:dm_learning}. 
Then, we propose to guide the policy learning by optimizing this estimate provided by a learned diffusion model, encouraging the policy to produce actions similar to expert actions, as discussed in~\mysecref{sec:dm_loss}.
Finally, in~\mysecref{sec:combining}, we introduce the framework that combines the BC loss and our proposed diffusion model loss, allowing for learning a policy that benefits from modeling both the \textit{conditional probability} and the \textit{joint probability} of expert behaviors.
An overview of our proposed framework is illustrated in~\myfig{fig:model}, and the algorithm is detailed in~\mysecref{sec:algorithm}.

\subsection{Behavioral Cloning Loss}
\label{sec:bc_learning}

The behavioral cloning (BC) model aims to imitate expert behaviors with supervision learning. BC learns to capture the conditional probability $p(a|s)$ of expert state-action pairs. A BC policy $\pi(a|s)$ learns by optimizing
\begin{equation}
\label{eq:bc_loss} 
\mathcal{L}_{\text{BC}} = \mathbb{E}_{(s, a) \sim D, \hat{a} \sim \pi(s)}[d(a, \hat{a})],
\end{equation}
where $d(\cdot, \cdot)$ denotes a distance measure between a pair of actions.
For example, we can adopt the mean-square error (MSE) loss ${||a - \hat{a}||}^2$ for most continuous control tasks.

\subsection{Learning a Diffusion Model and Guiding Policy Learning}
\label{sec:dm}
Instead of directly learning the conditional probability $p(a|s)$, this section discusses how to model the joint probability $p(s, a)$ of expert behaviors with a diffusion model in~\mysecref{sec:dm_learning} and presents how to leverage the learned diffusion model to guide policy learning in~\mysecref{sec:dm_loss}.

\subsubsection{Learning a Diffusion Model}
\label{sec:dm_learning}

We propose to model the joint probability of expert state-action pairs 
with a diffusion model $\phi$.
Specifically, we create a joint distribution by simply concatenating 
a state vector $s$ and an action vector $a$ from a state-action pair $(s, a)$.
To model such distribution by learning a denoising diffusion probabilistic model (DDPM)~\citep{ho2020ddpm}, we inject noise $\epsilon(n)$ into sampled state-action pairs, where $n$ indicates the number of steps of the Markov procedure, which can be viewed as a variable of the level of noise, and the total number of steps is notated as $N$. Then, we train the diffusion model $\phi$ to predict the injected noises by optimizing 
\begin{equation}
\begin{aligned}
\label{eq:diff_loss} 
    \mathcal{L}_{\text{diff}}(s, a, \phi) = \mathbb{E}_{n \sim N, (s, a) \sim D} \left[{||\hat{\epsilon}(s, a, n) - \epsilon(n)||}^2\right]& \\
    = \mathbb{E}_{n \sim N, (s, a) \sim D}\left[{||\phi(s, a, \epsilon(n)) - \epsilon(n)||}^2\right]&,
\end{aligned}
\end{equation}

where $\hat{\epsilon}$ is the noise predicted by the diffusion model $\phi$.
Once optimized, the diffusion model can \textit{recognize} the expert distribution by perfectly predicting the noise injected into state-action pairs sampled from the expert distribution.
On the other hand, predicting the noise injected into state-action pairs sampled from any other distribution should yield a higher loss value.
Therefore, we propose to view $\mathcal{L}_{\text{diff}}(s, a, \phi)$ as an estimate of how well the state-action pair $(s, a)$ fits the expert distribution that $\phi$ learns from and serve this estimate as a learning signal for the policy learning.

\subsubsection{Learning a Policy with Diffusion Model Loss}
\label{sec:dm_loss}

A diffusion model $\phi$ trained on an expert dataset can produce an estimate $\mathcal{L}_{\text{diff}}(s, a, \phi)$ indicating how well a state-action pair $(s, a)$ fits the expert distribution.
We propose to leverage this signal to guide a policy $\pi$ predicting actions $\hat{a}$ to imitate the expert.
Specifically, the policy $\pi$ learns by optimizing
\begin{equation}
\label{eq:agent_loss} 
\mathcal{L}_{\text{diff}}^{\text{agent}} = 
\mathcal{L}_{\text{diff}}(s, \hat{a}, \phi) = 
\mathbb{E}_{s \sim D, \hat{a} \sim \pi(s)}
\left[{||\hat{\epsilon}(s, \hat{a}, n) - \epsilon||}^2\right].
\end{equation}
Intuitively, the policy $\pi$ learns to predict actions $\hat{a}$ that are indistinguishable from the expert actions $a$ for the diffusion model conditioning on the same set of states. 
Note that the injected noise $\epsilon$ is drawn from a Gaussian distribution $\mathcal{G}(0, 1)$, and the diffusion step $n$ is drawn from the uniform distribution $\mathcal{U}(0, N)$. We omit these terms for simplicity in the equation and the following.

We hypothesize that learning a policy to optimize~\myeq{eq:agent_loss} can be unstable, especially for state-action pairs that are not well-modeled by the diffusion model, which yield a high value of $\mathcal{L}_{\text{diff}}$ even with expert state-action pairs.
Therefore, we propose to normalize the agent diffusion loss $\mathcal{L}_{\text{diff}}^{\text{agent}}$ with an expert diffusion loss $\mathcal{L}_{\text{diff}}^{\text{expert}}$, which can be computed with expert state-action pairs $(s, a)$ as follows:
\begin{equation}
\label{eq:expert_loss} 
\mathcal{L}_{\text{diff}}^{\text{expert}} = 
\mathcal{L}_{\text{diff}}(s, a, \phi) = 
\mathbb{E}_{(s, a) \sim D}
\left[{||\hat{\epsilon}(s, a, n) - \epsilon||}^2\right].
\end{equation}

We propose to optimize the diffusion model loss $\mathcal{L}_{\text{DM}}$ for the policy based on calculating the difference between the above agent and expert diffusion losses:

\begin{equation}
\label{eq:dm_loss} 
\mathcal{L}_{\text{DM}} = 
\mathbb{E}_{(s, a) \sim D, \hat{a} \sim \pi(s)}
\left[max\left(\mathcal{L}_{\text{diff}}^{\text{agent}}-\mathcal{L}_{\text{diff}}^{\text{expert}}, 0\right)\right].
\end{equation}

\subsection{Combining the Two Objectives}
\label{sec:combining}
Our goal is to learn a policy that benefits from both
modeling the conditional probability and the joint probability of expert behaviors. To this end, we propose to augment a BC policy, which optimizes the BC loss $L_{\text{BC}}$ in~\myeq{eq:bc_loss}, by combining $L_{\text{BC}}$ with the proposed diffusion model loss $L_{\text{DM}}$ in~\myeq{eq:dm_loss}.
By optimizing them together, we encourage the policy to predict actions that fit the expert joint probability captured by diffusion models.
To learn from both the BC loss and the diffusion model loss, we train the policy to optimize
\begin{equation}
\label{eq:total_loss} 
\mathcal{L}_{\text{total}} = 
\mathcal{L}_{\text{BC}} + 
\lambda
\mathcal{L}_{\text{DM}},
\end{equation}
where $\lambda$ is a coefficient that determines the importance of the diffusion model loss relative to the BC loss.
We include discussions on combining these two losses on both empirical (training progress) and theoretical (f-divergence) aspects in~\mysecref{sec:losses_relation}. Additionally, theoretical motivations for guiding policy learning with the diffusion model are shown in~\mysecref{sec:app_diffusion_model}.
\section{Experiments}
\label{sec:experiment}

We design experiments in various continuous control domains, including navigation, robot arm manipulation, dexterous manipulation, and locomotion, to compare our proposed framework (\method{}) to its variants and baselines.

\begin{figure*}[t]
    % \vspace{-10pt}
    \centering
    \begin{subfigure}[b]{0.16\textwidth}
    \centering
    \includegraphics[width=\textwidth, height=\textwidth]{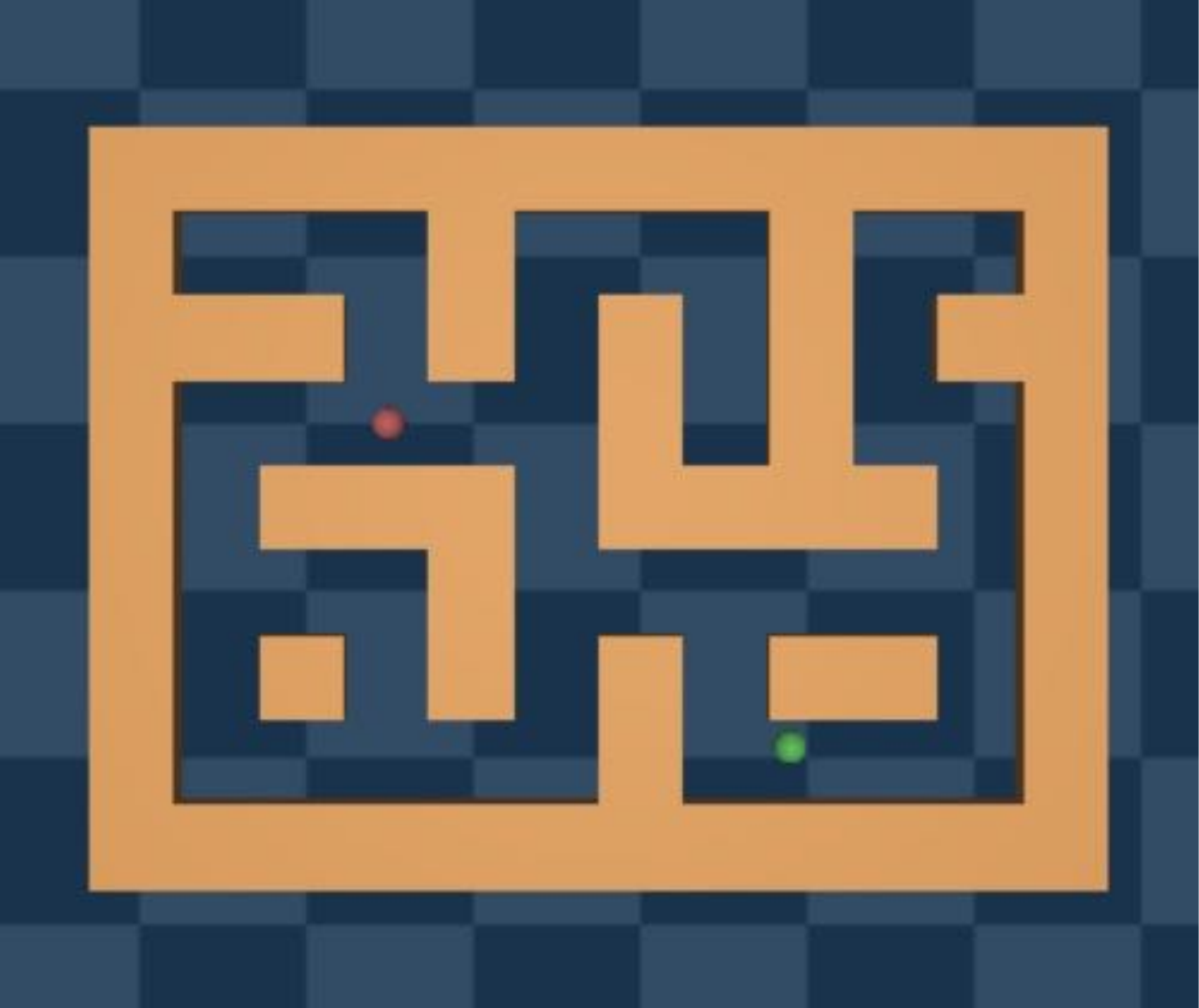}
    \caption{ \maze{}}
    \label{fig:env_maze}
    \end{subfigure}
    % \hfill
    \begin{subfigure}[b]{0.16\textwidth}
    \centering
    \includegraphics[width=\textwidth, height=\textwidth]{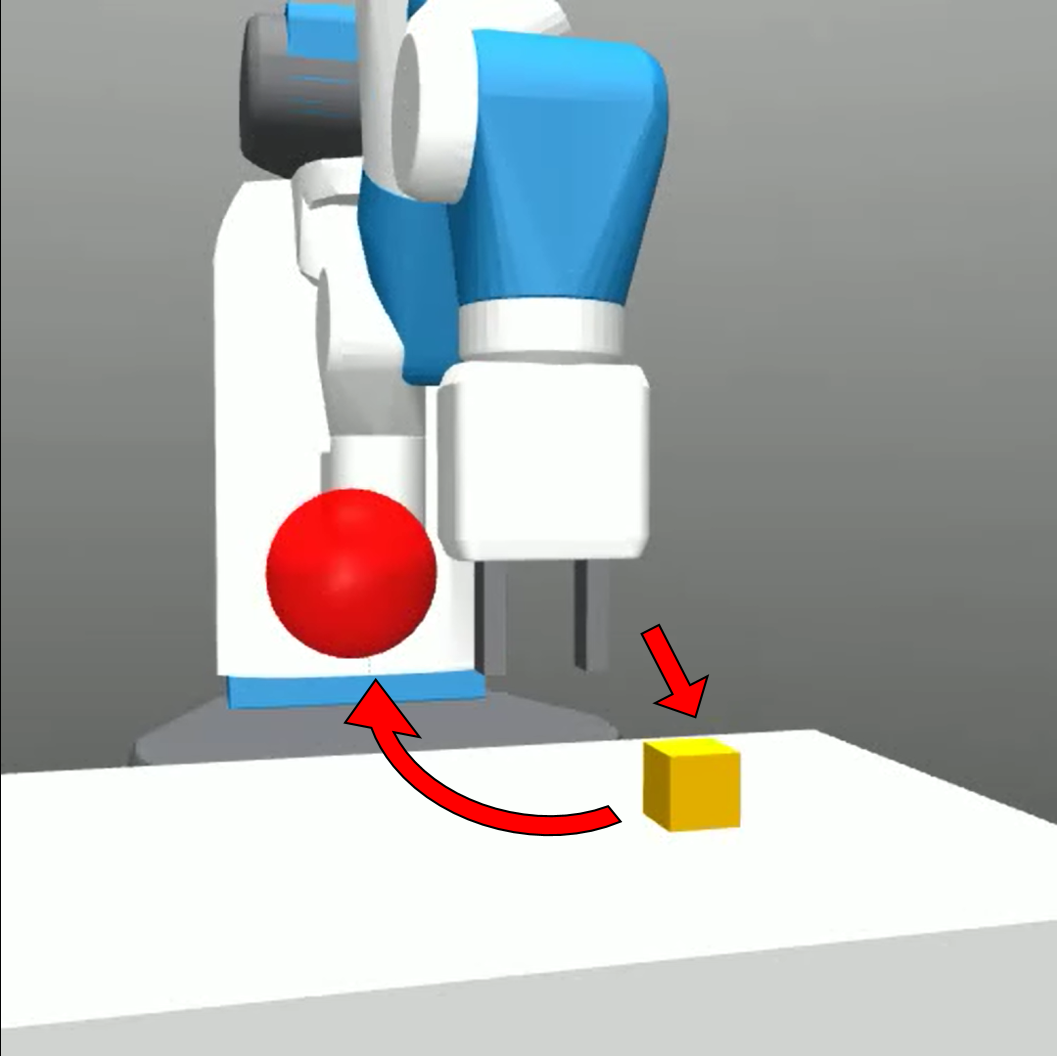}
    \caption{ \fetchpick{}}
    \label{fig:env_pick}    
    \end{subfigure}
    % \hfill
    \begin{subfigure}[b]{0.16\textwidth}
    \centering
    \includegraphics[width=\textwidth, height=\textwidth]{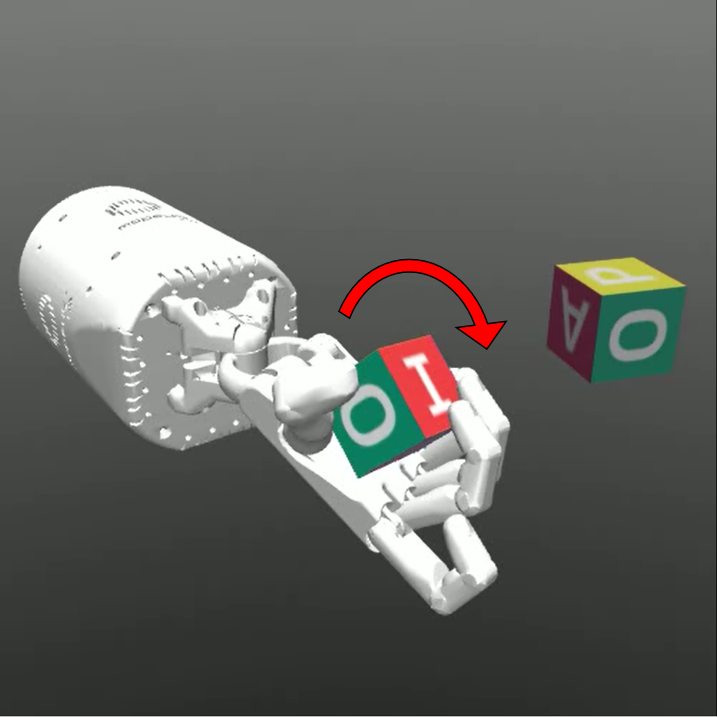}
    \caption{\textsc{HandRotate}}
    \label{fig:env_hand}
    \end{subfigure}    
    % \hfill
    \begin{subfigure}[b]{0.16\textwidth}
    \centering
    \includegraphics[width=\textwidth, height=\textwidth]{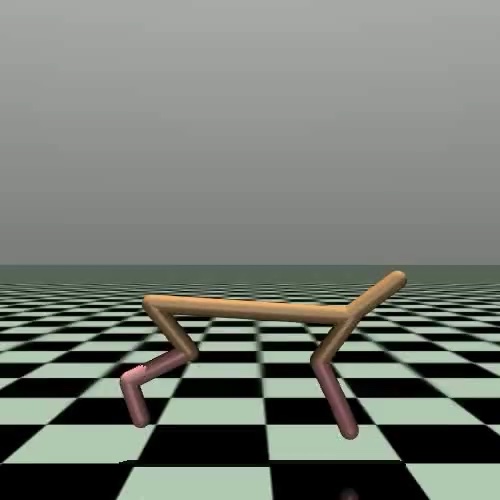}
     \caption{\textsc{Cheetah}}
    \label{fig:env_cheetah}    
    \end{subfigure}
    % \hfill
    \begin{subfigure}[b]{0.16\textwidth}
    \centering
    \includegraphics[width=\textwidth, height=\textwidth]{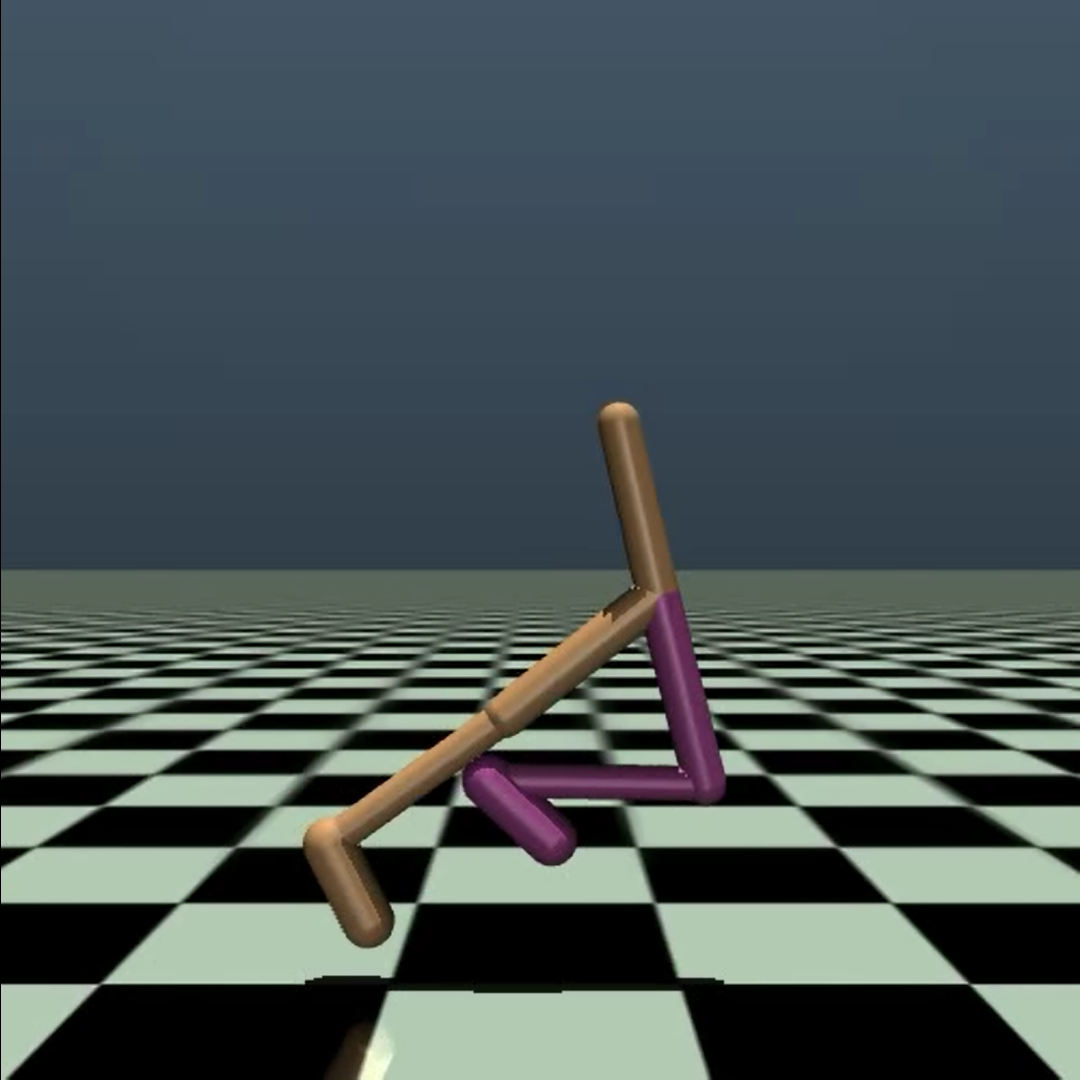}
    \caption{ \walker{}}
    \label{fig:env_walker}        
    \end{subfigure}
    % \hfill
    \begin{subfigure}[b]{0.16\textwidth}
    \centering
    \includegraphics[width=\textwidth, height=\textwidth]{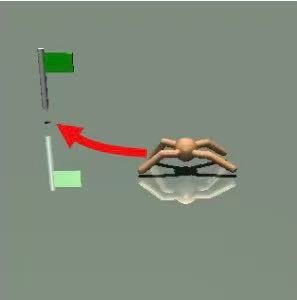}
    \caption{\antreach{}}
    \label{fig:env_ant}    
    \end{subfigure}
    % \hfill
    \vspace{-0.4cm}
    \caption[]{ 
    \textbf{Environments \& Tasks.}
    \textbf{(a) \maze{}}:
    A point-mass agent ({\color[RGB]{83, 144, 61}green}) in a 2D maze learns to navigate from its start location to a goal location ({\color[RGB]{207, 46, 42}red}).
    % by predicting its $x$ and $y$ velocity.
    \textbf{(b) \fetchpick{}}:
    The robot arm manipulation tasks employ a 7-DoF Fetch robotics arm to pick up an object ({\color[RGB]{224, 163, 38}yellow} cube) from the table and move it to a target location ({\color[RGB]{207, 46, 42}red}).
    \textbf{(c) \handrotate{}}:
    This dexterous manipulation task requires 
    a Shadow Dexterous Hand to in-hand rotate a block to a target orientation.
    % \textbf{(d) \cheetah{}}:
    % This locomotion task requires 
    % learning a bipedal walker policy
    % to walk as fast as possible.
    \textbf{(d)-(e) \cheetah{} and \walker{}}:
    These locomotion tasks require 
    learning agents
    to walk as fast as possible while maintaining their balance.
    \textbf{(f) \antreach{}}:
    This task combines locomotion and navigation, instructing an ant robot with four legs to reach a goal location while maintaining balance.}
    \label{fig:environment}
\end{figure*}

\subsection{Experimental Setup}
\label{sec:small_experiment}

This section describes the environments, tasks, and expert demonstrations used for learning and evaluation. More details can be found in~\mysecref{sec:app_exp_task}.

\myparagraph{Navigation}
To evaluate our method on a navigation task,
we choose \maze{}, a maze environment proposed in~\citet{fu2020d4rl} (maze2d-medium-v2), 
as illustrated in \myfig{fig:env_maze}.
This task features a point-mass agent in a 2D maze learning to navigate from its start location to a goal location by iteratively predicting its $x$ and $y$ acceleration. 
The agent's beginning and final locations are chosen randomly. We collect 100 demonstrations with 18,525 transitions using a controller. 

\myparagraph{Robot Arm Manipulation}
We evaluate our method in \fetchpick{}, a robot arm manipulation domain with a 7-DoF Fetch task, 
as illustrated in~\myfig{fig:env_pick}.
\fetchpick{} requires picking up an object from the table and lifting it to a target location.
We use the demonstrations, consisting of 10k transitions (303 trajectories), provided by~\citet{goalprox} for these tasks.

\myparagraph{Dexterous Manipulation} 
In \handrotate{}, we further evaluate our method on a challenging environment proposed in~\citet{plappert2018multi}, where a 24-DoF Shadow Dexterous Hand learns to in-hand rotate a block to a target orientation, as illustrated in \myfig{fig:env_hand}. 
This environment has a state space (68D) and action space (20D), which is high dimensional compared to the commonly-used environments in IL. We collected 10k transitions (515 trajectories) from a SAC~\citep{haarnoja18b} expert policy trained for 10M environment steps.

\myparagraph{Locomotion} 
For locomotion, we leverage the \cheetah{} and \walker{}~\citep{1606.01540} environments. Both \cheetah{} and \walker{} require a bipedal agent (with different structures) to travel as fast as possible while maintaining its balance, as illustrated in \myfig{fig:env_cheetah} and \myfig{fig:env_walker}, respectively. 
We use the demonstrations provided by~\citet{walkerdemo}, which contains 5 trajectories with 5k state-action pairs for both the \cheetah{} and \walker{} environments.

\myparagraph{Locomotion + Navigation}
We further explore our method on the challenging \antreach{} environment.
In the environment, the quadruped ant aims to reach a randomly generated target located along the boundary of a semicircle centered around the ant, as illustrated in \myfig{fig:env_ant}.
\antreach{} environment combines the properties of locomotion and goal-directed navigation tasks, which require robot controlling and path planning to reach the goal.
We use the demonstrations provided by~\citet{goalprox}, which contains 500 trajectories with 25k state-action pairs in \antreach{}.

\subsection{Baselines}
\label{sec:exp_baselines}
This work focuses on imitation learning problem \textit{without} environment interactions. Therefore, approaches that require environmental interactions, such as GAIL-based methods, are not applicable.
Instead, we extensively compared our proposed method to state-of-the-art imitation learning methods that do not require interaction with the environment, including Implicit BC~\citep{florence2022implicit} and Diffusion Policy~\citep{chi2023diffusionpolicy, reuss2023goal}. 

\begin{itemize}
    \item \textbf{BC} learns to imitate an expert by modeling the conditional probability $p(a|s)$ of the expert behaviors via optimizing the BC loss $\mathcal{L}_{\text{BC}}$ in~\myeq{eq:bc_loss}.
    
    \item \textbf{Implicit BC (IBC)} \citep{florence2022implicit} models expert state-action pairs with an energy-based model. For inference, we implement the derivative-free optimization algorithm proposed in IBC, which samples actions iteratively and selects the desired action according to the predicted energies.
    
    \item \textbf{Diffusion policy} refers to the methods that learn a conditional diffusion model as a policy~\citep{chi2023diffusionpolicy, reuss2023goal}.
    Specifically, we implement this baseline based on~\citet{pearce2022imitating}.
    We include this baseline to analyze the effectiveness of using diffusion models as a policy or as a learning objective (ours).
\end{itemize}

\begin{table}
\centering
\normalsize
\caption[]{\textbf{Multimodality of Environments}.
We evaluate the multimodality of expert trajectories of each environment by measuring if the states in the same cluster share actions from the same clusters. The ratio ranges from 0.1 to 1, indicating whether states within the same cluster perform actions that are either randomly distributed (1/10) or consistently identical (1/1), respectively. 
}
\vspace{0.2cm}
\scalebox{0.85}{
\begin{tabular}{@{}cc@{}}\toprule
\textbf{Environment} & Majority Ratio \\
\cmidrule{1-2}
\maze{} & 0.184 \\ 
\fetchpick{} & 0.604 \\ 
\handrotate{} & 0.331 \\ 
\cheetah{} & 0.594 \\ 
\walker{} & 0.582 \\ 
\antreach{} & 0.511 \\ 
\bottomrule
\end{tabular}
}
\label{table:multimodality}
\end{table}
\begin{table*}
\centering
\caption[]{\textbf{Experimental Result.}
We report the mean and the standard deviation of success rate (\maze{}, \fetchpick{}, \handrotate{}, \antreach{}) and return (\cheetah{}, \walker{}),
evaluated over three random seeds.
Our proposed method (\method{}) outperforms or performs competitively against the best baseline over all environments.
}
\vspace{0.2cm}
%\ra{1.3}
\scalebox{0.85}{\begin{tabular}{@{}ccccccc@{}}\toprule
\textbf{Method} & 
\maze{} &
\fetchpick{} &
\handrotate{} &
\cheetah{} &
\walker{} &
\antreach{} \\
\cmidrule{1-7}
BC & 92.1\% $\pm$ 3.6\% 
& 91.6\% $\pm$ 5.8\% 
& 57.5\% $\pm$ 4.7\% 
& \textbf{4873.3} $\pm$ 69.7
& 6954.4 $\pm$ 73.5
& 56.2\% $\pm$ 4.9\% \\
Implicit BC & 78.3\% $\pm$ 6.0\% 
& 69.4\% $\pm$ 7.3\% 
& 13.8\% $\pm$ 3.7\% 
& 1563.6 $\pm$ 486.8
& 839.8 $\pm$ 104.2
& 23.7\% $\pm$ 4.9\% \\
Diffusion Policy & \textbf{95.5}\% $\pm$ 1.9\% 
& 83.9\% $\pm$ 3.4\% 
& \textbf{61.7}\% $\pm$ 4.1\% 
& 4650.3 $\pm$ 59.9
& 6479.1 $\pm$ 238.6
& 61.8\% $\pm$ 4.0\% \\
\method{}  (Ours) & \textbf{95.4}\% $\pm$ 1.7\% 
& \textbf{97.5}\% $\pm$ 1.9\% 
& \textbf{60.1}\% $\pm$ 4.4\% 
& \textbf{4909.5} $\pm$ 73.0
& \textbf{7034.6} $\pm$ 33.7
& \textbf{70.1}\% $\pm$ 4.9\% \\
\bottomrule
\end{tabular}}
\label{table:main}
\end{table*}
\subsection{Multimodality of Environments}
\label{sec:exp_multimodal}
In this section, we aim to quantitatively evaluate the multimodality of expert trajectories of each environment we use in the paper.
IBC and DP are well-known for their ability to handle multimodal data, and understanding the multimodality in each environment can help us better compare with these baselines.
In imitation learning, multimodality may arise from either the nature of the task, \eg different goals with arbitrary orders, or the expert demonstrations, \eg achieving the same goal with various paths.
For each task, we create 10 clusters of states and 10 clusters of actions from expert demonstrations. Then, we measure if the states in the same cluster share actions from the same clusters. Specifically, we calculate the major action class for each state cluster and compute the ratio of states with the class. The ratio ranges from 0.1 to 1, indicating whether states within the same cluster perform actions that are either randomly distributed (1/10) or consistently identical (1/1), respectively. 

The results of all the tasks are reported in~\mytable{table:multimodality}.
We observe that robot arm manipulation (\fetchpick{}) and locomotion (\cheetah{} and \walker{}) tasks result in higher majority ratios, which indicates that the expert behaviors are more unimodal.
On the other hand, navigation (\maze{}) and dexterous manipulation (\handrotate{}) tasks result in lower majority ratios, which means the demonstrations contain more multimodal paths for similar goals and \antreach{} results in an intermediate majority ratio since it is a combination of navigation and locomotion.

\begin{table*}
\centering
\large
\caption[]{\textbf{Generalization Experiments in \fetchpick{}}.
We report the performance of our proposed framework \method{} and the baselines regarding the mean and the standard deviation of the success rate 
with different levels of noise injected into
the initial state and goal locations in \fetchpick{},
evaluated over three random seeds.
}
\vspace{0.2cm}
\scalebox{0.85}{\begin{tabular}{@{}ccccccc@{}}\toprule
\multirow{2}{*}{\textbf{Method}} & \multicolumn{5}{c}{\textbf{Noise Level}} \\
& 1 & 1.25 & 1.5 & 1.75 & 2 \\
\cmidrule{1-6}
BC & 92.4\% $\pm$ 8.5\% 
& 91.6\% $\pm$ 5.8\% 
& 85.5\% $\pm$ 6.3\% 
& 77.6\% $\pm$ 7.1\%
& \textbf{67.4}\% $\pm$ 8.2\% \\
Implicit BC & 83.1\% $\pm$ 3.1\%
& 69.4\% $\pm$ 7.3\% 
& 51.6\% $\pm$ 4.2\% 
& 36.5\% $\pm$ 4.7\%
& 23.6\% $\pm$ 3.0\% \\
Diffusion Policy & 90.0\% $\pm$ 3.5\% 
& 83.9\% $\pm$ 3.4\% 
& 72.3\% $\pm$ 6.8\% 
& 64.1\% $\pm$ 7.1\%
& 58.2\% $\pm$ 8.2\%\\
\method{}  (Ours) & \textbf{99.5}\% $\pm$ 0.5\% 
& \textbf{97.5}\% $\pm$ 1.9\%
& \textbf{91.5}\% $\pm$ 3.3\%
& \textbf{83.3}\% $\pm$ 4.8\%
& \textbf{73.5}\% $\pm$ 6.8\%\\
\bottomrule
\end{tabular}}
\label{table:generalization_pick}
\end{table*}
\subsection{Experimental Results}
\label{sec:exp_result}
We report the experimental results 
in terms of success rate (\maze{}, \fetchpick{}, \handrotate{}, and \antreach{}), 
and return (\cheetah{} and \walker{}) in~\mytable{table:main}.
The details of model architecture can be found in~\mysecref{sec:app_model}.
Training and evaluation details can be found in~\mysecref{sec:app_training_evaluation}.
Additional analysis and experimental results can be found in~\mysecref{sec:app_ablation},~\mysecref{sec:app_add_noise}, and~\mysecref{sec:app_data}.

\myparagraph{Overall Task Performance}
In navigation (\maze{}) and dexterous manipulation (\handrotate{}) tasks, our \method{} performs competitively, i.e., within a standard deviation, against the Diffusion Policy and outperforms the other baselines.
As discussed in~\mysecref{sec:exp_multimodal}, these tasks require the agent to learn from multimodal demonstrations of various behaviors.
Diffusion policy has shown promising performance for capturing multi-modality distribution, while our \method{} can also generalize well with the guidance of the diffusion models, so both methods achieve satisfactory results.

In locomotion tasks, i.e., \cheetah{} and \walker{}, our \method{} outperforms Diffusion Policy and performs competitively against the simple BC baseline.
We hypothesize that this is because locomotion tasks with sufficient expert demonstrations and little randomness do not require generalization during inference, which results in lower majority scores as shown in~\mysecref{sec:exp_multimodal}.
The agent can simply follow the closed-loop progress of the expert demonstrations, resulting in both BC and \method{} performing similarly to the expert demonstrations. On the other hand, the Diffusion Policy is designed for modeling multimodal behaviors and therefore, performs inferior results on single-mode locomotion tasks. For \antreach{} task, which combines locomotion and navigation, our method outperforms all the baselines. 

In summary, our proposed \method{} is able to perform superior results across all tasks, which verifies the effectiveness of combining conditional and joint distribution modeling. 

\myparagraph{Inference Efficiency}
To evaluate the inference efficiency,
we measure and report the number of evaluation episodes per second ($\uparrow$) for Implicit BC (9.92), Diffusion Policy (1.38), and \method{} (\textbf{30.79}) on an NVIDIA RTX 3080 Ti GPU in \maze{}.
As a result of modeling the conditional probability $p(a|s)$,
\method{} and BC can directly map states to actions during inference. 
In contrast, Implicit BC samples and optimizes actions, while Diffusion Policy iteratively denoises sampled noises,
which are both time-consuming.
This verifies the efficiency of modeling the conditional probability.

\myparagraph{Action Space Dimension}
The Implicit BC baseline requires time-consuming action sampling and optimization during inference, and such a procedure may not scale well to high-dimensional action spaces.
Our Implicit BC baseline with a derivative-free optimizer struggles in \cheetah{}, \walker{}, and \handrotate{} environments, whose action dimensions are 6, 6, and 20, respectively.
This is consistent with~\citet{florence2022implicit}, 
which reports that the optimizer failed to solve tasks with an action dimension larger than 5.
In contrast, our proposed \method{} can handle high-dimensional action spaces.

\subsection{Generalization Experiments in \fetchpick{}}
\label{sec:app_exp}

This section further investigates the generalization capabilities of the policies learned by our proposed framework and the baselines.
To this end, we evaluate the policies by injecting different noise levels to both the initial state and goal location in \fetchpick{}.
Specifically, we parameterize the noise by scaling the 2D sampling regions for the block and goal locations in both environments. 
We expect all the methods to perform worse with higher noise levels, while the performance drop of the methods with better generalization ability is less significant.
In this experiment, we set the coefficient $\lambda$ of \method{} to $0.1$ in \fetchpick{}.
The results are presented in \mytable{table:generalization_pick} for \fetchpick{}.

\myparagraph{Overall Performance} Our proposed framework \method{} consistently outperforms all the baselines with different noise levels, 
indicating the superiority of 
\method{} when different levels of generalization are required.

\myparagraph{Performance Drop with Increased Noise Level} 
In \fetchpick{}, 
\method{} experiences a performance drop of $26.1\%$ when the noise level increase from $1$ to $2$.
However, BC and Implicit BC demonstrate a performance drop of $27.0\%$ and $71.6\%$, respectively.
Notably, Diffusion Policy initially performs poorly at a noise level of $1$ but demonstrates its robustness with a performance drop of only $35.3\%$ when the noise level increases to $2$. 
This shows that our proposed framework generalizes better and exhibits greater robustness to noise compared to the baselines.

\begin{figure}[t]
    % \centering
    % \begin{subfigure}[b]{0.26\textwidth}
    % \centering
    % \includegraphics[width=\textwidth, height=\textwidth]{figure/toy/toy.001.pdf}
    % \caption{}
    % \label{fig:toy_generalization}
    % \end{subfigure}
    % \hspace{0.45cm}
    \begin{subfigure}[b]{0.235\textwidth}
    \centering
    \includegraphics[width=\textwidth, height=\textwidth]{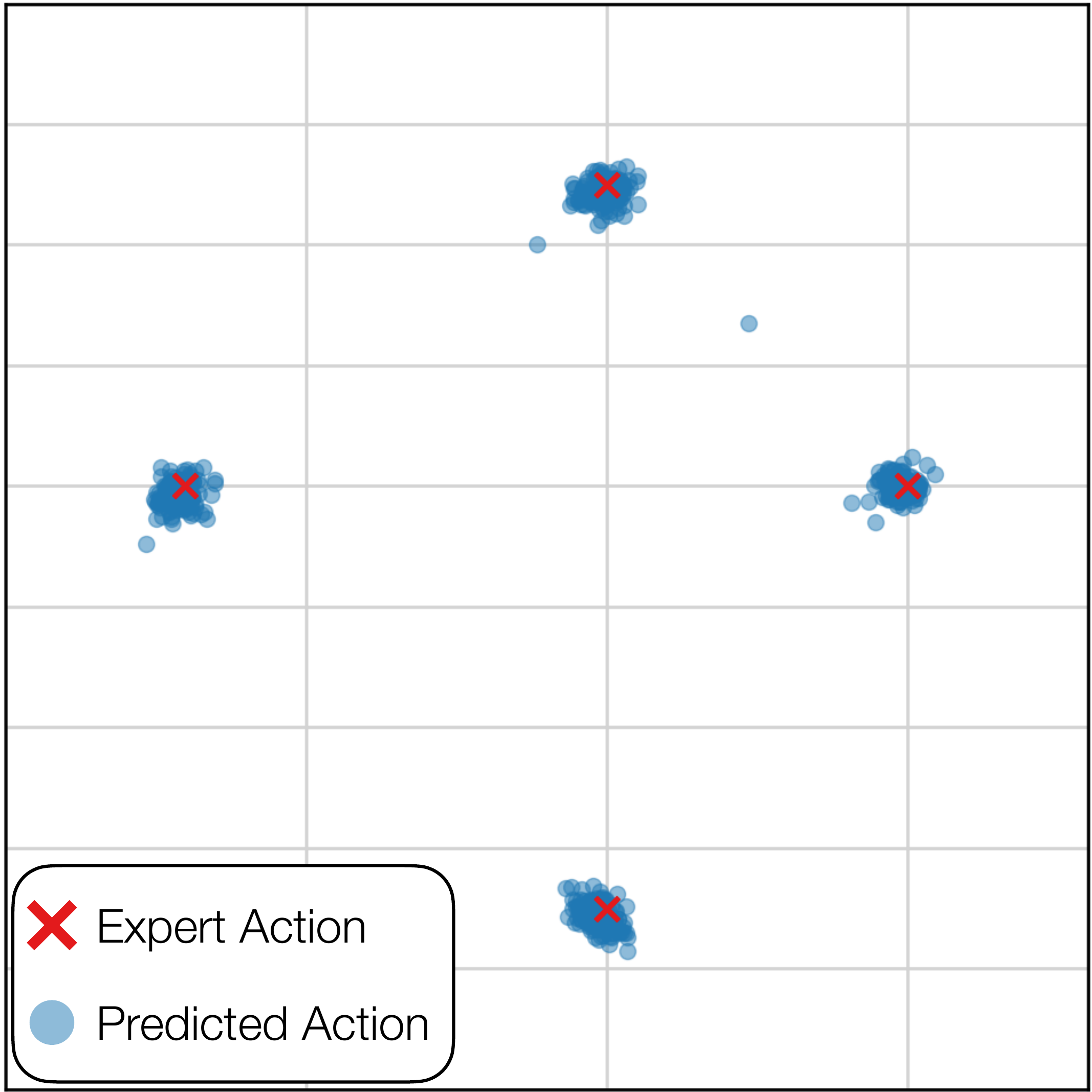}
    \caption{Sampled actions}
    \label{fig:toy_manifold_overfitting_action}
    \end{subfigure}
    \hspace{0.05cm}
    \begin{subfigure}[b]{0.235\textwidth}
    \centering
    \includegraphics[width=\textwidth, height=\textwidth]{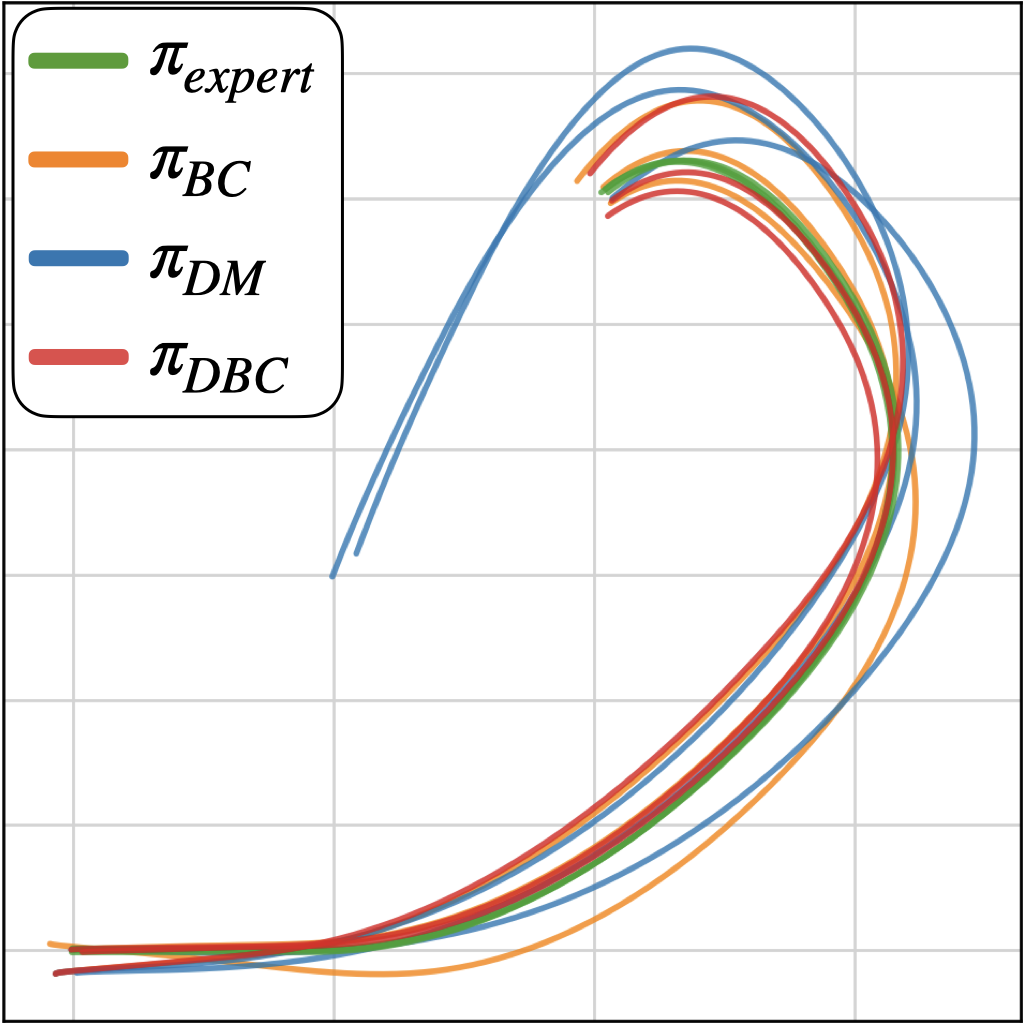}
    \caption{Trajectories}
    \label{fig:toy_manifold_overfitting}
    \end{subfigure} 
    % \hfill
    % \begin{subfigure}[b]{0.16\textwidth}
    % % \label{}
    % \centering
    % \includegraphics[width=\textwidth, height=\textwidth]{figure/env_figure/maze.pdf}
    % \caption{DM}
    % \end{subfigure}  
    \vspace{-0.4cm}
    \caption[]{
    % \textbf{Comparing Modeling Conditional Probability and Joint Probability.}  
    % \textbf{(a) Generalization.}
    % We collect expert trajectories from a PPO policy learning to navigate to goals sampled from the {\color[RGB]{83, 144, 61}green} regions. 
    % Then, we learn a policy $\pi_{BC}$ to optimize $\mathcal{L}_{\text{BC}}$, 
    % and another policy $\pi_{DM}$ to optimize $\mathcal{L}_{\text{DM}}$ with a diffusion model trained on the expert distribution.
    % We evaluate the two policies by sampling goals from the {\color[RGB]{207, 46, 42}red} regions, 
    % which requires the ability to generalize. 
    % $\pi_{BC}$ ({\color[RGB]{238, 121, 53}orange}) struggles at generalizing to unseen goals, 
    % whereas $\pi_{DM}$ ({\color[RGB]{45, 108, 163}blue}) can generalize (\ie extrapolate) to some extent.
    \textbf{Manifold overfitting Experiments.}
    (a) We collect the {\color[RGB]{83, 144, 61}green} spiral trajectories from a script policy, 
    whose actions are visualized as {\color[RGB]{207, 46, 42}red} crosses.
    (b) We train and evaluate $\pi_{BC}$, $\pi_{DM}$ and $\pi_{DBC}$ using the demonstrations from the script policy.
    The trajectories of $\pi_{BC}$ ({\color[RGB]{238, 121, 53}orange}) and $\pi_{DBC}$ ({\color[RGB]{207, 46, 42}red}) can closely follow the expert trajectories ({\color[RGB]{83, 144, 61}green}),
    while the trajectories of $\pi_{DM}$ ({\color[RGB]{45, 108, 163}blue}) deviates from expert's. 
    This is because the diffusion model struggles at modeling such expert action distribution with a lower intrinsic dimension,
    which can be observed from incorrectly predicted actions ({\color[RGB]{45, 108, 163}blue} dots) produced by the diffusion model.
    % \textbf{Conclusion.} These two experiment's results verify our motivation to complement modeling the joint probability (\ie diffusion model) with modeling the conditional probability (\ie BC).
    }
    \label{fig:toy}
\end{figure}

\subsection{Manifold Overfitting Experiments}
\label{sec:conditional_vs_joint}

This section aims to empirically examine if modeling joint probabilities is difficult when observed high-dimensional data points lie on a low-dimensional manifold (\ie, manifold overfitting).
We employ a point maze environment implemented with~\citet{fu2020d4rl} and collect trajectories from a script policy that executes actions $(0.5, 0)$, $(0, 0.5)$, $(-0.7, 0)$, and $(0, -0.7)$ (red crosses in~\myfig{fig:toy_manifold_overfitting_action}), each for 40 consecutive time steps, resulting the green spiral trajectories visualized in~\myfig{fig:toy_manifold_overfitting}.

Given these expert demonstrations, we learn a policy $\pi_{BC}$ to optimize~\myeq{eq:bc_loss}, a policy $\pi_{DM}$ to optimize~\myeq{eq:dm_loss} with a diffusion model trained on the expert distribution, and a policy $\pi_{DBC}$ to optimize the combined objective~\myeq{eq:total_loss}.
\myfig{fig:toy_manifold_overfitting_action} shows that the diffusion model struggles at modeling such expert action distribution with a lower intrinsic dimension.
As a result, \myfig{fig:toy_manifold_overfitting}
show that the trajectories of $\pi_{DM}$ ({\color[RGB]{45, 108, 163}blue}) deviates from the expert trajectories ({\color[RGB]{83, 144, 61}green}) as the diffusion model cannot provide effective loss.
On the other hand, the trajectories of $\pi_{BC}$ ({\color[RGB]{238, 121, 53}orange}) and $\pi_{DBC}$ ({\color[RGB]{207, 46, 42}red}) are both able to closely follow the expert's and result in a superior success rate.
This verifies our motivation to complement modeling the joint probability with modeling the conditional probability (\ie BC).
\section{Discussion}
\label{sec:conclusion}

We propose an imitation learning framework that benefits from
modeling both the conditional probability
$p(a|s)$ and the joint probability $p(s, a)$ of the expert distribution.
Our proposed \methodFull{} (\method{}) employs a diffusion model trained to model expert behaviors and learns a policy to optimize both the BC loss and our proposed diffusion model loss.
Specifically, the BC loss captures the conditional probability $p(a|s)$ from expert state-action pairs, which directly guides the policy to replicate the expert's action. 
On the other hand, the diffusion model loss models the joint distribution of expert state-action pairs $p(s, a)$, which provides an evaluation of how well the predicted action aligned with the expert distribution. 
\method{} outperforms baselines or achieves competitive performance in various continuous control tasks in navigation, robot arm manipulation, dexterous manipulation, and locomotion. 
We design additional experiments to verify the limitations of modeling either the conditional probability or the joint probability of the expert distribution as well as compare different generative models.
Ablation studies investigate the effect of hyperparameters and justify the effectiveness of our design choices.
Despite its encouraging results, our proposed framework is designed to learn from expert trajectories without interacting with environments and cannot learn from agent trajectories.
Extending our method to incorporate agent data can potentially allow for improvement when interacting environments is possible.

\section*{Acknowledgement}
\label{sec:ack}
This work was supported by the National Science and Technology Council, Taiwan (NSTC 112-2222-E-002-006-).
Shao-Hua Sun was supported by the Yushan Fellow Program by the Ministry of Education, Taiwan. 

\section*{Impact Statement}
This work proposes \methodFull{}, a novel imitation learning framework that aims to increase the ability of autonomous learning agents (\eg robots, game AI agents) to acquire skills by imitating demonstrations provided by experts (\eg humans). 
However, it is crucial to acknowledge that our proposed framework, by design, inherits any biases exhibited by the expert demonstrators. These biases can manifest as sub-optimal, unsafe, or even discriminatory behaviors. 
To address this concern, ongoing research endeavors to mitigate bias and promote fairness in machine learning hold promise in alleviating these issues.
Moreover, research works that enhance learning agents' ability to imitate experts, such as this work, can pose a threat to job security.
Nevertheless, in sum, we firmly believe that our proposed framework can offer tremendous advantages in terms of enhancing the quality of human life and automating laborious, arduous, or perilous tasks that pose risks to humans, which far outweigh the challenges and potential issues.

\bibliography{main}
\bibliographystyle{icml2024}

%%%%%%%%%%%%%%%%%%%%%%%%%%%%%%%%%%%%%%%%%%%%%%%%%%%%%%%%%%%%%%%%%%%%%%%%%%%%%%%
%%%%%%%%%%%%%%%%%%%%%%%%%%%%%%%%%%%%%%%%%%%%%%%%%%%%%%%%%%%%%%%%%%%%%%%%%%%%%%%
% APPENDIX
%%%%%%%%%%%%%%%%%%%%%%%%%%%%%%%%%%%%%%%%%%%%%%%%%%%%%%%%%%%%%%%%%%%%%%%%%%%%%%%
%%%%%%%%%%%%%%%%%%%%%%%%%%%%%%%%%%%%%%%%%%%%%%%%%%%%%%%%%%%%%%%%%%%%%%%%%%%%%%%
\newpage
\onecolumn
\appendix
\section*{Appendix}

% \begingroup
% \hypersetup{pdfborder={0 0 0}}

% \part{} % Start the appendix part
% \parttoc % Insert the appendix TOC
% % \listoftables
% % \listoffigures

% \endgroup

\section{Algorithm}
\label{sec:algorithm}
Our proposed framework \method{} is detailed in~\myalgo{alg:algo}.
The algorithm consists of two parts.
(1) \textbf{Learning a diffusion model}: The diffusion model $\phi$ learns to model the distribution of concatenated state-action pairs sampled from the demonstration dataset $D$. It learns to reverse the diffusion process (\ie denoise) by optimizing $\mathcal{L}_\text{diff}$.
(2) \textbf{Learning a policy with the learned diffusion model}: We propose a diffusion model objective $\mathcal{L}_{\text{DM}}$ for policy learning and jointly optimize it with the BC objective $\mathcal{L}_{\text{BC}}$. Specifically, $\mathcal{L}_{\text{DM}}$ is computed based on processing a sampled state-action pair $(s, a)$ and a state-action pair $(s, \hat{a})$ with the action $\hat{a}$ predicted by the policy $\pi$ with $\mathcal{L}_\text{diff}$.
\begin{algorithm}[H]
\caption{\methodFull{} (\method{})} 
\label{alg:algo}
\textbf{Input}: Expert's Demonstration Dataset $D$ \\ % and Diffusion Model $\phi$ \\
\textbf{Output}: Policy $\pi$.
\begin{algorithmic}[1]
\STATE {\color{gray} // Learning a diffusion model $\phi$}
\STATE Randomly initialize a diffusion model $\phi$ 
\FOR{each diffusion model iteration}
    \STATE Sample $(s, a)$ from $D$
    \STATE Sample noise level $n$ from $\{0, ..., N\}$
    \STATE Update $\phi$ using $L_{\text{diff}}$ from~\myeq{eq:diff_loss}
\ENDFOR
\STATE {\color{gray}// Learning a policy $\pi$  with the learned diffusion model $\phi$}
\STATE Randomly initialize a policy $\pi$ 
\FOR{each policy iteration}
    \STATE Sample $(s, a)$ from $D$
    \STATE Predict an action $\hat{a}$ using 
    % the policy 
    $\pi$ from 
    %the sampled state 
    $s$: $\hat{a} \sim \pi(s)$
    \STATE Compute the BC loss $L_{\text{BC}}$ using~\myeq{eq:bc_loss} 
    \STATE Sample noise level $n$ from $\{0, ..., N\}$
    \STATE Compute the agent diffusion loss $L_{\text{diff}}^{\text{agent}}$ with $(s, \hat{a})$ using~\myeq{eq:agent_loss} 
    \STATE Compute the expert diffusion loss  $L_{\text{diff}}^{\text{expert}}$ with $(s, a)$ using~\myeq{eq:expert_loss}
    \STATE Compute the diffusion model loss $L_{\text{DM}}$ using~\myeq{eq:dm_loss} 
    \STATE Update $\pi$ using the total loss $L_{\text{total}}$ from~\myeq{eq:total_loss}
\ENDFOR
\STATE \textbf{return} $\pi$
\end{algorithmic}
\end{algorithm}

\begin{wrapfigure}[12]{R}{0.32\textwidth}
    \vspace{-0.4cm}
    \centering
    \includegraphics[width=\linewidth]{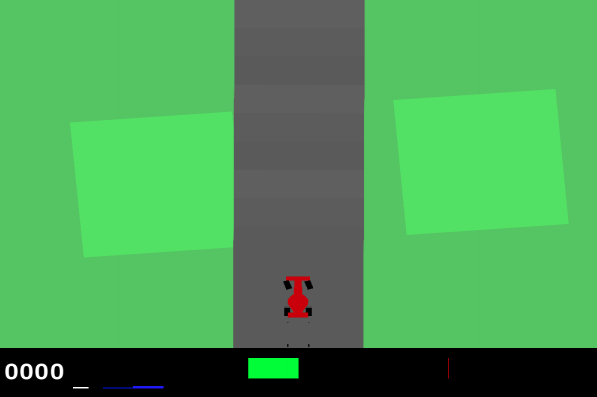}
    \vspace{-0.7cm}
    \caption{\textbf{\carracing{}.} This task features controlling a car navigating a track with image-based observations.}
    \label{fig:env_car}
\end{wrapfigure}
\section{Additional Experiments on Image-Based Environment}
\label{sec:app_image_based}
We have additionally conducted experiments on an image-based environment, in which each state is represented as an RGB image.
Specifically, we adopt a racing game environment \carracing{} that aims to learn a policy to control a car to navigate a track, as illustrated in~\myfig{fig:env_car}.
We employ $5$ trajectories with $1826$ transitions from a PPO expert policy.
The state represents two consecutive top-down $96 \times 96$ RGB images capturing the car and the track and is converted to $64 \times 64$ grayscale images to reduce the computation cost.
The action is a three-dimensional vector that indicates values of steering, acceleration, and breaking.

We compare our \method{} to a BC baseline in this task.
BC achieves an average return of $766.4$ with a standard deviation of $26.3$, outperformed by our proposed \method{} with an average return of $791.5$ and a standard deviation of $27.7$.
This highlights that the proposed method can improve the performance of BC in an image-based environment by incorporating the proposed diffusion model loss~\myeq{eq:dm_loss}, aligning with the conclusion drawn from the tasks with vectorized states presented in our main paper.

\section{Ablation Study}
\label{sec:app_ablation}
\subsection{Comparing Different Generative Models}
\label{sec:app_generative_models}

Our proposed framework employs a diffusion model (DM) to model the joint probability of expert state-action pairs and utilizes it to guide policy learning.
To justify our choice, we explore using other popular generative models to replace the diffusion model in \maze{} and \fetchpick{}.  We consider energy-based models (EBMs)~\citep{du2019implicit, song2021train}, variational autoencoder (VAEs)~\citep{kingma2013auto}, and generative adversarial networks (GANs)~\citep{goodfellow2014generative}.
Each generative model learns to model expert state-action pairs.
To guide policy learning, given a predicted state-action pair $(s, \hat{a})$ we use the estimated energy of an EBM, the reconstruction error of a VAE, and the discriminator output of a GAN to optimize a policy with or without the BC loss. 

\mytable{table:gm} and \mytable{table:gm_pick} compares using different generative models to model the expert distribution and guide policy learning on \maze{} and \fetchpick{} environments, respectively. 
All the generative model-guide policies can be improved by adding the BC loss, justifying our motivation to complement modeling the joint probability with modeling the conditional probability.
With or without the BC loss, the diffusion model-augmented policy achieves the best performance compared to other generative models.
Specifically, DM outperforms the second-best baseline GAN by 24.8\% improvement without BC and by 2.4\% with BC on \maze{}. The results verify our choice of the generative model.
Training details of learning the generative models and how to utilize them to guide policy learning can be found in~\mysecref{sec:app_gm}.

\begin{table}[t]
\begin{minipage}{0.49\linewidth}
% \large
\centering
\caption[]{\textbf{Comparing Generative Models in \maze{}.}
We compare using different generative models to model the expert distribution and guide policy learning in \maze{}. With or without the BC loss, the diffusion model-augmented policy achieves the best performance compared to other generative models.}
\vspace{0.2cm}
\begin{tabular}{@{}ccc@{}}\toprule
\textbf{Method} & without BC & with BC \\
\cmidrule{1-3}
BC & N/A & 92.1\% $\pm$ 3.6\%\\
EBM &  20.3\% $\pm$ 11.8\% & 92.5\% $\pm$ 3.0\%\\
VAE &  53.1\% $\pm$ 8.7\% & 92.7\% $\pm$ 2.7\%\\
GAN &  54.8\% $\pm$ 4.4\% & 93.0\% $\pm$ 3.5\%\\
DM  &  \textbf{79.6}\% $\pm$ 9.6\% & \textbf{95.4}\% $\pm$ 1.7\%\\
\bottomrule
\end{tabular}
\label{table:gm}
\end{minipage}
\hspace{0.02\linewidth}
\begin{minipage}{0.49\linewidth}
\centering
\caption[]{\textbf{Comparing Generative Models in \fetchpick{}.} 
We compare using different generative models to model the expert distribution and guide policy learning in \fetchpick{}. With or without the BC loss, the diffusion model-augmented policy achieves the best performance compared to other generative models.
}
\vspace{0.2cm}
%\ra{1.3}
\begin{tabular}{@{}ccc@{}}\toprule
\textbf{Method} & without BC & with BC \\
\cmidrule{1-3}
BC & N/A
& 91.6\% $\pm$ 5.8\%\\
EBM &  5.5\% $\pm$ 7.0\%
& 90.7\% $\pm$ 5.9\%\\
VAE &  0.7\% $\pm$ 0.8\%
& 94.6\% $\pm$ 2.1\%\\
GAN &  \textbf{41.8}\% $\pm$ 24.9\%
& 90.3\% $\pm$ 4.3\%\\
DM  &  14.2\% $\pm$ 16.2\%
& \textbf{97.5}\% $\pm$ 1.9\%\\
\bottomrule
\end{tabular}
\label{table:gm_pick}
\label{figure:coeff}
\end{minipage}
\end{table}

\subsection{Effect of the Diffusion Model Loss Coefficient $\lambda$}
\label{sec:coefficient_ablation}

We examine the impact of varying the coefficient of
the diffusion model loss $\lambda$ in~\myeq{eq:total_loss} in \fetchpick{}.
The result presented in~\myfig{fig:coeff_pick} shows that $\lambda=0.5$ yields the best performance of $97.5\%$. A higher or lower $\lambda$ leads to worse performance. For instance, when $\lambda$ is 0 (only BC), the success rate is $91.7\%$, and the performance drops to $51.46\%$ when $\lambda$ is 10. This result demonstrates that modeling the conditional probability ($\mathcal{L}_{\text{BC}}$) and the joint probability ($\mathcal{L}_{\text{DM}}$) can complement each other.

Our guideline for selecting the coefficient $\lambda$ is to ensure that the behavior cloning loss $L_{\text{BC}}$ and the diffusion model loss $L_{\text{DM}}$ are approximately of the same order of magnitude. As shown in~\mytable{table:hyperparameter}, optimal $\lambda$ varies from task to task since the loss scale also varies.
However, it is relatively easy to determine $\lambda$, and the performance of \method{} is reasonably robust to different $\lambda$ as long as the orders of magnitude of the two losses are balanced.

\subsection{Effect of the Normalization Term}
\label{sec:ablation_agent_loss}
We aim to investigate whether normalizing the diffusion model loss $\mathcal{L}_{\text{DM}}$ with the expert diffusion model loss $\mathcal{L}_{\text{diff}}^{\text{expert}}$ yields improved performance.
We train a variant of \method{} where only $\mathcal{L}_{\text{diff}}^{\text{agent}}$ in~\myeq{eq:agent_loss} instead of $\mathcal{L}_{\text{DM}}$ in~\myeq{eq:dm_loss} is used to augment BC.
For instance, the unnormalized variant performs worse than \method{} in the \maze{} environment, where the average success rate is $94\%$ and $95\%$, respectively. This justifies the effectiveness of the proposed normalization term $\mathcal{L}_{\text{diff}}^{\text{expert}}$ in $\mathcal{L}_{\text{DM}}$.
We find consistent results in all of the environments except \antreach{}, and comprehensive results can be found in~\mytable{table:expert_normalization}.

\begin{table}[t]
\begin{minipage}{0.48\linewidth}
\centering
\includegraphics[width=0.8\linewidth]{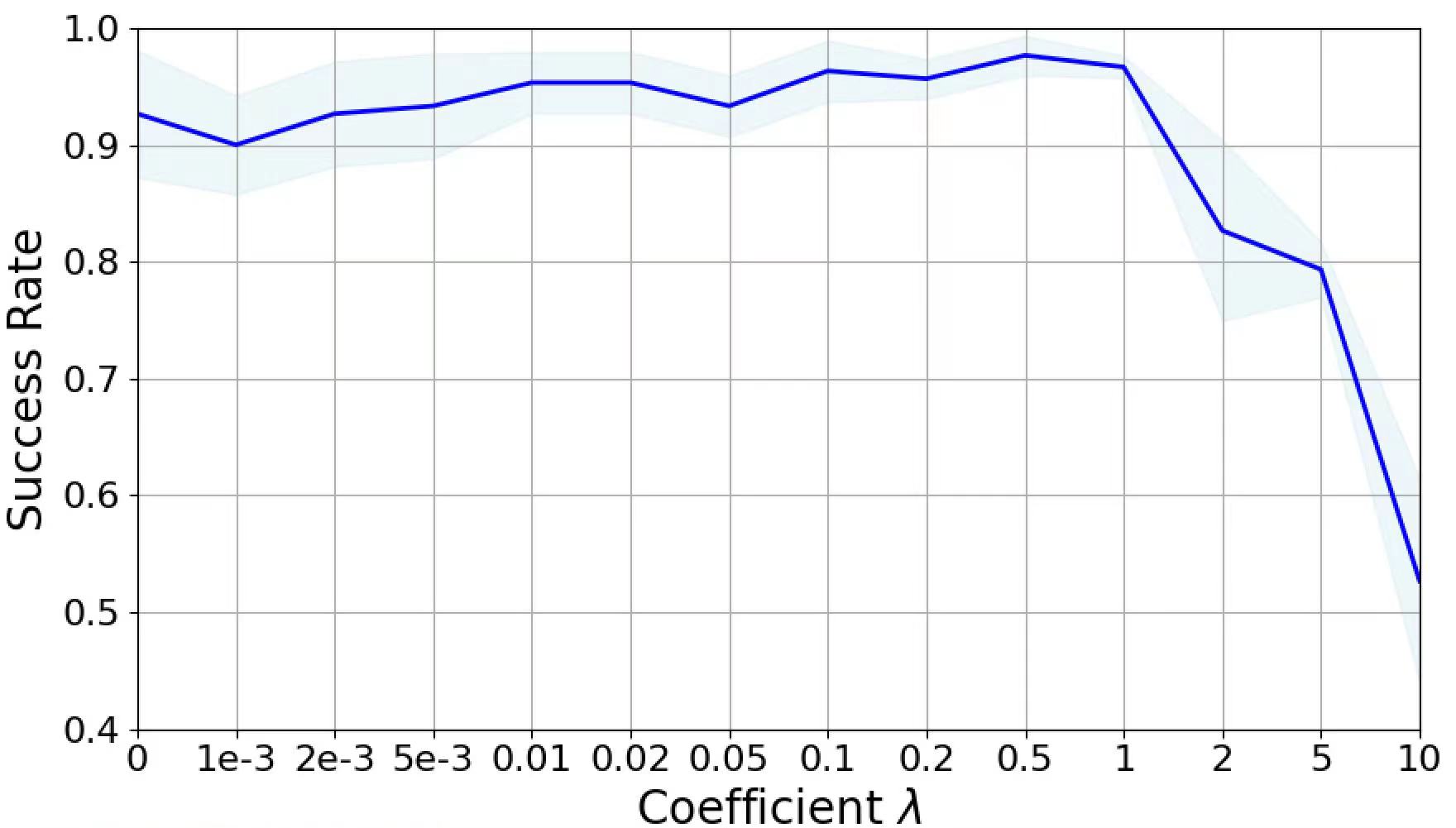}
\captionof{figure}[]{\textbf{Effect of Coefficient $\lambda$ for $L_{DM}$.} We experiment with different values of $\lambda$ in \fetchpick{}, each evaluated over three random seeds.}
\label{fig:coeff_pick}
\end{minipage}
\hspace{0.3cm}
\begin{minipage}{0.49\linewidth}
\centering
\caption[Effect of Normalization Term]{\textbf{Effect of Normalization Term.}
To investigate the effectiveness of the normalization term, we evaluate a variant of \method{} where only $\mathcal{L}_{\text{diff}}^{\text{agent}}$ in~\myeq{eq:agent_loss} instead of $\mathcal{L}_{\text{DM}}$ in~\myeq{eq:dm_loss} is used. 
}
\scalebox{0.9}{\begin{tabular}{@{}ccccc@{}}\toprule
\textbf{Environment} & $\mathcal{L}_{\text{diff}}^{\text{agent}}$ &  $\mathcal{L}_{\text{DM}}$\\
\cmidrule{1-3}
\maze{} & 94.7\% $\pm$ 1.9\%
& 95.4\% $\pm$ 1.7\%\\
\fetchpick{} & 96.6\% $\pm$ 1.7\%
& 96.9\% $\pm$ 1.7\%\\
\handrotate{} & 59.4\% $\pm$ 2.1\%
& 60.1\% $\pm$ 4.4\%\\
\antreach{} & 70.1\% $\pm$ 4.9\%
& 70.0\% $\pm$ 5.0\%\\
\cheetah{} & 4821.4 $\pm$ 124.0
& 4909.5 $\pm$ 73.0\\
\walker{} & 6976.4 $\pm$ 76.1
& 7034.6 $\pm$ 33.7\\
\bottomrule
\end{tabular}}
\label{table:expert_normalization}
\end{minipage}
\end{table}

\section{Relationships between $\mathcal{L}_{\text{BC}}$ and $\mathcal{L}_{\text{DM}}$}
\label{sec:losses_relation}
\subsection{Training Progress}
We notice that when the learned policy is optimal, \ie $\pi = \pi^{E}$, both objectives converge to 0 
despite that $\mathcal{L}_{\text{BC}}$ models the conditional probability while $\mathcal{L}_{\text{DM}}$ models the joint probability.
Therefore, we examine the roles of the above two objectives during the training procedure.

\begin{figure}[t]
    \centering
    \includegraphics[width=0.3\textwidth]{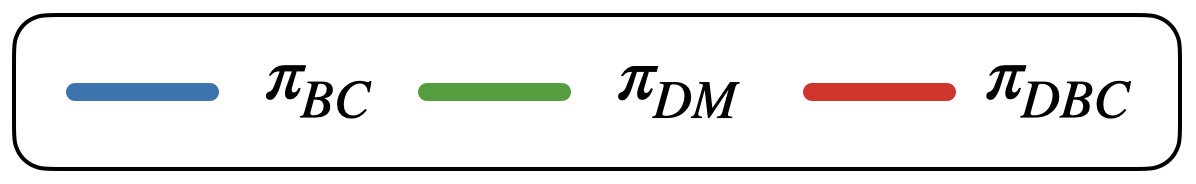}
    \vspace{0.5cm}
    
    \begin{subfigure}{0.35\textwidth}
            \centering
            \includegraphics[width=\textwidth]{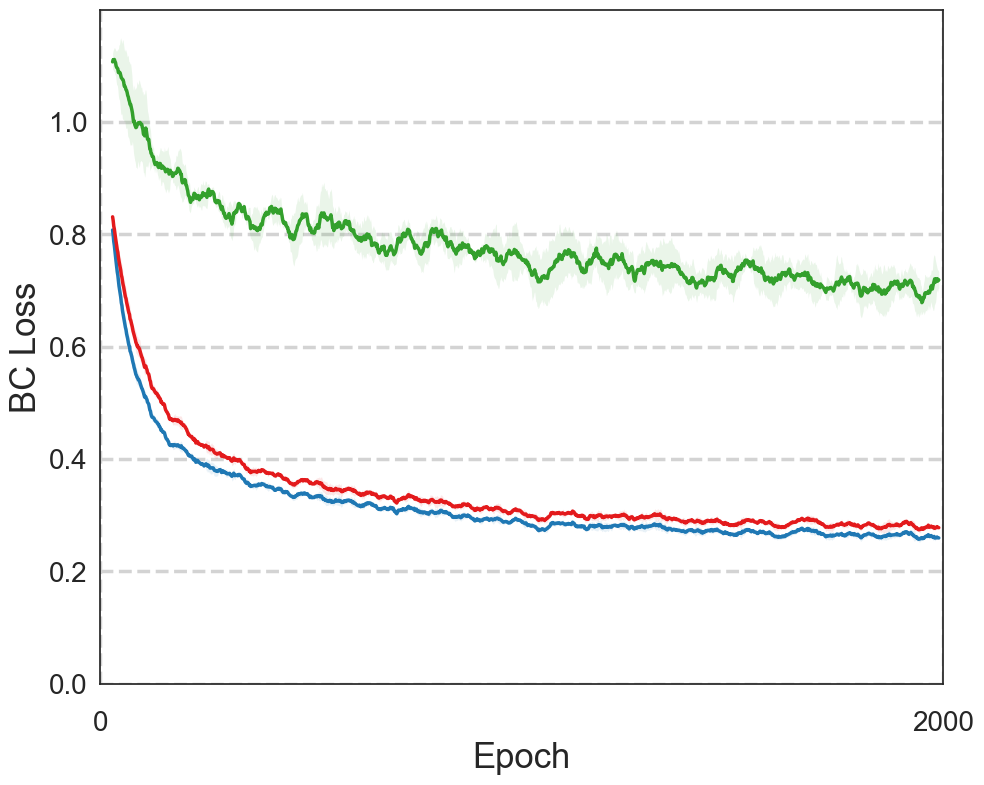}
            \caption{Training curve of $\mathcal{L}_{\text{BC}}$}
            \label{fig:bc_loss}
    \end{subfigure}
    \begin{subfigure}{0.35\textwidth}
            \centering
            \includegraphics[width=\textwidth]{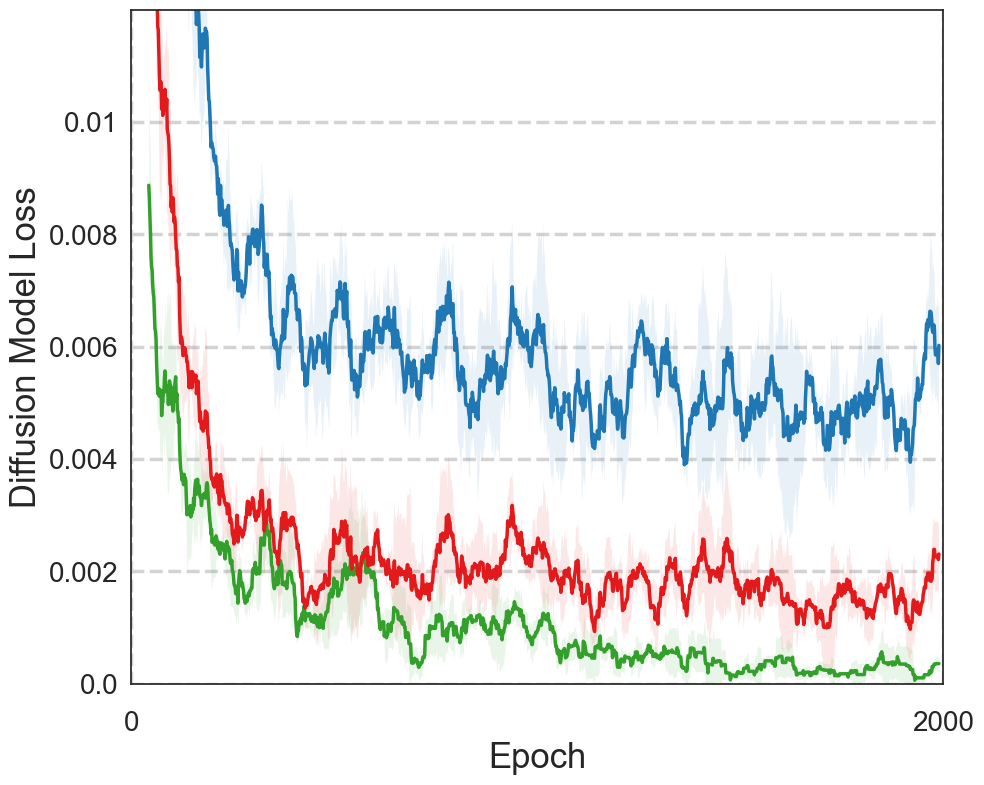}
            \caption{Training curve of $\mathcal{L}_{\text{DM}}$}
            \label{fig:dm_loss}
    \end{subfigure}
    
    \caption{\textbf{Compatibility of $\mathcal{L}_{\text{BC}}$ and $\mathcal{L}_{\text{DM}}$.} We report the training progress of three policies $\pi_{BC}$, $\pi_{DM}$, and $\pi_{DBC}$ that are updated with $\mathcal{L}_{\text{BC}}$, $\mathcal{L}_{\text{DM}}$, and both objectives, respectively. Our proposed method can effectively optimize both $\mathcal{L}_{\text{BC}}$ and $\mathcal{L}_{\text{DM}}$, demonstrating the compatibility of the two losses.
    }
    \label{fig:loss_curve}
\end{figure}

As~\myfig{fig:loss_curve} shown, we train three policies on \maze{} environment with $L_{BC}$, $L_{DM}$, and both objectives. The derived policies are referred to $\pi_{BC}$, $\pi_{DM}$, and $\pi_{DBC}$, respectively.
We observe that while optimizing one loss can also reduce the other loss to some extent ($\pi_{BC}$ and $\pi_{DM}$), optimizing the combination leads to favorable convergence for both objectives ($\pi_{DBC}$).
Therefore, considering both objectives, that is, considering both the conditional and the joint probability, is beneficial for policy learning and leads the learned policy $\pi$ closer to the optimal one $\pi^{E}$.
The above observation is also supported by the quantitative results presented in both~\mytable{table:main} and~\mytable{table:gm} in the subsequent section.

% \definecolor{myblue}{RGB}{18, 110, 168}
% \begin{figure}[t]
% % \begin{wrapfigure}[19]{R}{6.875cm} % 0.55textwidth
%     \vspace{-0.25cm}
%     \centering
%     \includegraphics[width=2.0625cm] {figure/loss_curve/loss_legend.jpeg} % 0.3\textwidth
%     \vspace{0.05cm}
%     \\
%     \begin{subfigure}[]%{3.375cm} % c
%         \centering
%         \includegraphics[width=3.375cm] % \textwidth
%         {figure/loss_curve/bc_loss.png}
%         \subcaption{Training Progress of $\mathcal{L}_{\text{BC}}$}
%         % \caption{Training Progress of $\mathcal{L}_{\text{BC}}$}
%         \label{fig:bc_loss}
%     \end{subfigure}
%         \begin{subfigure}[]%{3.375cm} %0.27\textwidth
%         \centering
%         \includegraphics[width=3.375cm] %\textwidth
%         {figure/loss_curve/dm_loss.png}
%         \subcaption{Training Progress of $\mathcal{L}_{\text{DM}}$}
%         % \caption{Training Progress of $\mathcal{L}_{\text{DM}}$}
%         \label{fig:dm_loss}
%     \end{subfigure}

%     \caption[]{\textbf{Compatibility of $\mathcal{L}_{\text{BC}}$ and $\mathcal{L}_{\text{DM}}$.} We report the training progress of three policies $\pi_{BC}$, $\pi_{DM}$, and $\pi_{DBC}$ that are updated with $\mathcal{L}_{\text{BC}}$, $\mathcal{L}_{\text{DM}}$, and both objectives, respectively. Our proposed method can effectively optimize both $\mathcal{L}_{\text{BC}}$ and $\mathcal{L}_{\text{DM}}$, demonstrating the compatibility of the two losses.
%     }
%     \label{fig:loss_curve}
% % \end{wrapfigure}
% \end{figure}
% \usepackage{subcaption}
% \renewcommand\thesubfigure{(\alph{subfigure})}

\definecolor{myblue}{RGB}{18, 110, 168}

\subsection{F-Divergence}
\label{sec:f_divergence}
To provide theoretical motivation for our method, we show that optimizing the BC loss can be approximated to optimizing the forward Kullback–Leibler (KL) divergence while optimizing the diffusion model loss can be approximated to optimizing the reverse KL divergence.

As shown by~\citet{ke2021imitation}, a policy minimizing the KL divergence of the distribution of an expert policy $\pi^{E}$ can be represented as $\hat{\pi} = - \mathbb{E}_{s \sim \rho_{\pi^{E'}}, a \sim \pi^{E}} [log(\pi(a|s))]$, which is equivalent to the BC objective with the use of a cross-entropy loss.
In this work, we adopt Gaussian policies, which are widely used in continuous control tasks, to predict an action from a state.
According to~\citet{goodfellow2016deep}, assuming the target distribution is a unit Gaussian with a fixed variance $\sigma$, the cross-entropy loss, i.e., minimization of the conditional negative log-likelihood (NLL), can be reparameterized to a mean squared error (MSE) with the following form:
\begin{equation}
\Sigma^{m}_{i=1}-\log p(y_i|x_i; \Theta) = 
m\log \sigma + {\frac{m}{2}}\log(2\pi)+\Sigma^{m}_{i=1}{\frac{||\hat{y}_i - y_i||^2}{2{\sigma}^2}},
\end{equation}
where $x_i$ is the input, $y_i$ is the label, $\hat{y}_i$ is the prediction from model $\Theta$, and $m$ is the number of samples.
Accordingly, minimizing $\mathcal{L}_{\text{BC}}$ (\myeq{eq:bc_loss}) is equal to minimizing the forward KL divergence between the expert distribution and the agent one.

Next, we show how the diffusion model loss can be approximated to optimize the reverse KL divergence in the following.
As shown in~\citet{ho2020ddpm}, the noise prediction objective can optimize the variational bound on negative log probability $\mathbb{E}_{(s,a) \sim D} [-\log \rho_{\pi^{E}} (s,a)]$.
In~\myeq{eq:agent_loss}, our diffusion loss $\mathcal{L}_{\text{diff}}^{\text{agent}}$ takes the states $s$ sampled from the dataset $D$ and actions $\hat{a}$ predicted by the policy $\pi$.
Therefore, given a pre-trained diffusion model that captures the approximation of expert distribution $\rho_{\pi^{E'}}$, 
we can do the following derivation for the variational bound:
\begin{equation}
\begin{aligned}
&\mathbb{E}_{s \sim D, \hat{a} \sim \pi} [-\log \rho_{\pi^{E'}}(s,\hat{a})] \\
& = \iint [ -\rho_{\pi}(s,\hat{a}) \log \rho_{\pi^{E'}}(s,\hat{a})] \,ds\,d\hat{a} \\
& = \iint [-\rho_{\pi}(s,\hat{a}) \log \rho_{\pi^{E'}}(s,\hat{a}) + \rho_{\pi}(s,\hat{a}) \log \rho_{\pi}(s,\hat{a}) - \rho_{\pi}(s,\hat{a}) \log \rho_{\pi}(s,\hat{a})] \,ds\,d\hat{a} \\ 
& = \iint \rho_{\pi}(s,\hat{a}) [\log\rho_{\pi}(s,\hat{a}) - \log\rho_{\pi^{E'}}(s,\hat{a})] \,ds\,d\hat{a} + \iint -\rho_{\pi}(s,\hat{a}) \log \rho_{\pi}(s,\hat{a}) \,ds\,d\hat{a} \\
& = \mathrm{D_{RKL}}(\rho_{\pi^{E'}}(s,\hat{a}), \rho_{\pi}(s,\hat{a})) + \mathcal{H}(\rho_{\pi}) \\
& \geq \mathrm{D_{RKL}} (\rho_{\pi^{E'}}(s,\hat{a}), \rho_{\pi}(s,\hat{a})),
\end{aligned}
\end{equation}

where $\rho_{\pi^{E'}}$ and $\rho_{\pi}$ represent the distribution of the estimated expert and the agent state-action pairs respectively, $\mathcal{H}(\rho_{\pi})$ represents the entropy of $\rho_{\pi}$, and $\mathrm{D_{RKL}} (\rho_{\pi^{E'}}, \rho_{\pi})$ represents the reverse KL divergence of $\rho_{\pi^{E'}}$ and $\rho_{\pi}$. As a result, we can minimize the reverse KL divergence between the estimated expert distribution and the agent distribution by optimizing the diffusion loss.

As shown by~\citet{ke2021imitation}, the forward KL divergence promotes mode coverage at the cost of occasionally generating poor samples. In contrast, optimizing the reverse KL helps generate high-quality samples at the cost of sacrificing modes. 
Therefore, the above derivation indicates that $\mathcal{L}_{\text{BC}}$ and $\mathcal{L}_{\text{DM}}$ exhibit complementary attributes, which motivates us to combine the BC loss and the proposed diffusion model loss.

\section{Alleviating Manifold Overfitting by Noise Injection}
\label{sec:app_add_noise}

In section \mysecref{sec:conditional_vs_joint}, we show that while our diffusion model loss can enhance the generalization ability of the derived policy, the diffusion models may suffer from manifold overfitting during training and, therefore, need to cooperate with the BC objective.
Another branch of machine learning research dealing with overfitting problems is noise injection. As shown in~\citet{feng2021understanding}, noise injection regularization has shown promising results that resolve the overfitting problem on image generation tasks. 
In this section, we evaluate if noise injection can resolve the manifold overfitting directly.

\subsection{Modeling Expert Distribution}
We first verify if noisy injection can help diffusion models capture the expert distribution of the spiral dataset, where the diffusion models fail as shown in~\mysecref{sec:conditional_vs_joint}.
We extensively evaluate diffusion models trained with various levels of noise added to the expert actions. Then, we calculate the average MSE distance between expert actions and the reconstruction of the trained diffusion models, which indicates how well diffusion models capture the expert distribution.
We report the result in~\mytable{table:noise_level}.

We observe that applying a noise level of less than 0.02 results in similar MSE distances compared to the result without noise injection (0.0213).
The above result indicates that noise injection does not bring an advantage to expert distribution modeling regarding the MSE distance, and the discrepancy between the learned and expert distributions still exists.

\begin{table}
\centering
\caption[Modeling Noisy Expert distribution]{\textbf{Modeling Noisy Expert distribution.} Expert distribution modeling with diffusion models trained with different noise levels.}

\scalebox{1}{\begin{tabular}{@{}cccccccc@{}}\toprule
\textbf{Noise level} & 
0 &
0.002 &
0.005 &
0.01 &
0.02 &
0.05 &
0.1 \\

\cmidrule{1-8}
MSE Distance 
& 0.0213
& 0.0217
& 0.0248
& 0.0218
& 0.0235 
& 0.0330
& 0.0507
\\
\bottomrule
\end{tabular}}
\label{table:noise_level}
\end{table}

\subsection{Guide Policy Learning}
In order to examine if the noise-injected diffusion model is better guidance for policy, we further investigate the performance of using the learned diffusion models to guide policy learning.
Specifically, we train policies to optimize the diffusion model loss $\mathcal{L}_{\text{DM}}$ provided by either the diffusion model learning from a noise level of 0 or the diffusion model learning from a noise level of 0.01, dubbed $\pi_{\text{DM-0.01}}$. 

\begin{figure}[t]
    \centering
    
    \begin{minipage}[b]{0.23\textwidth}
        \centering
        \includegraphics[width=\textwidth, height=\textwidth]{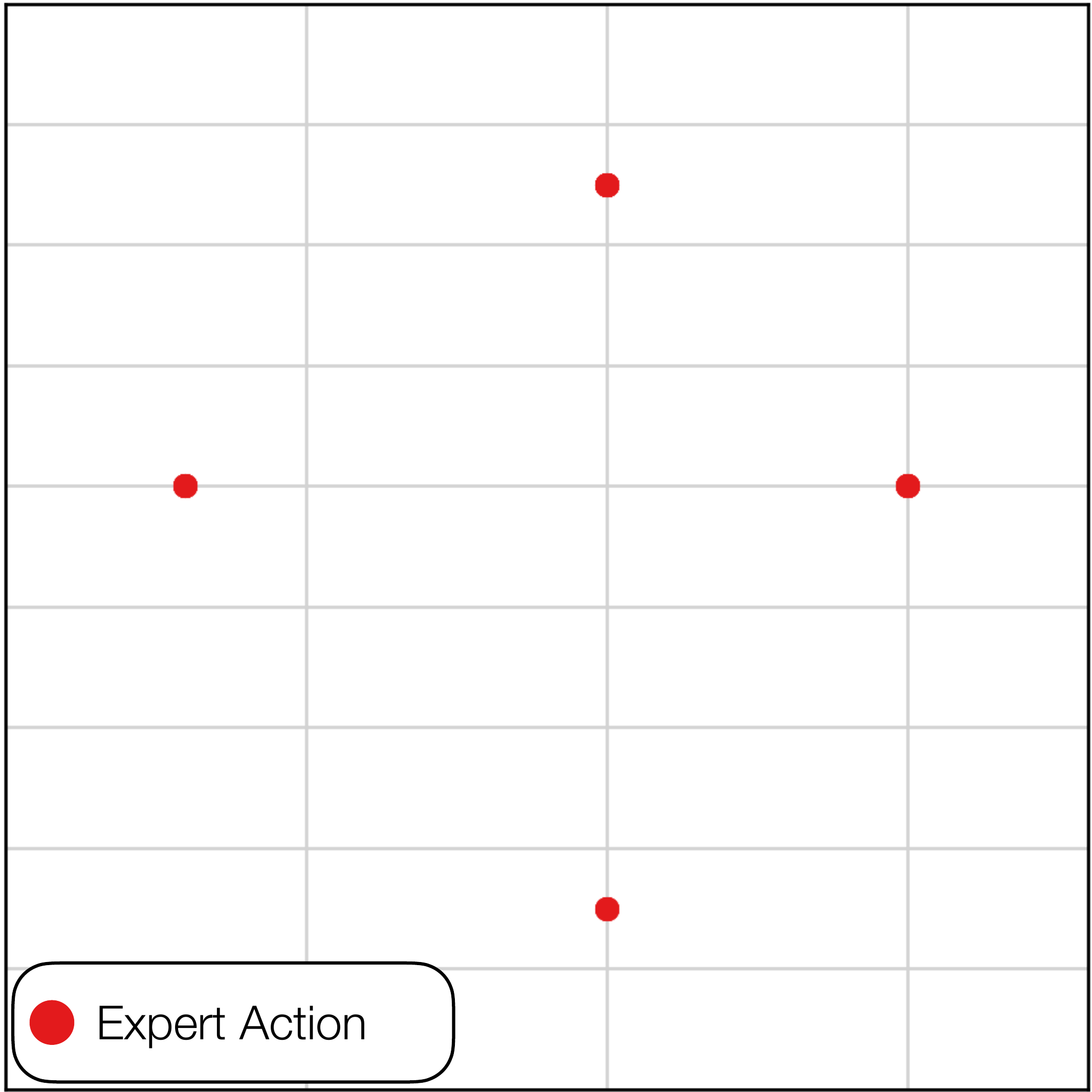}
        \caption{}
    \end{minipage}
    \hspace{0.15cm}
    \begin{minipage}[b]{0.23\textwidth}
        \centering
        \includegraphics[width=\textwidth, height=\textwidth]{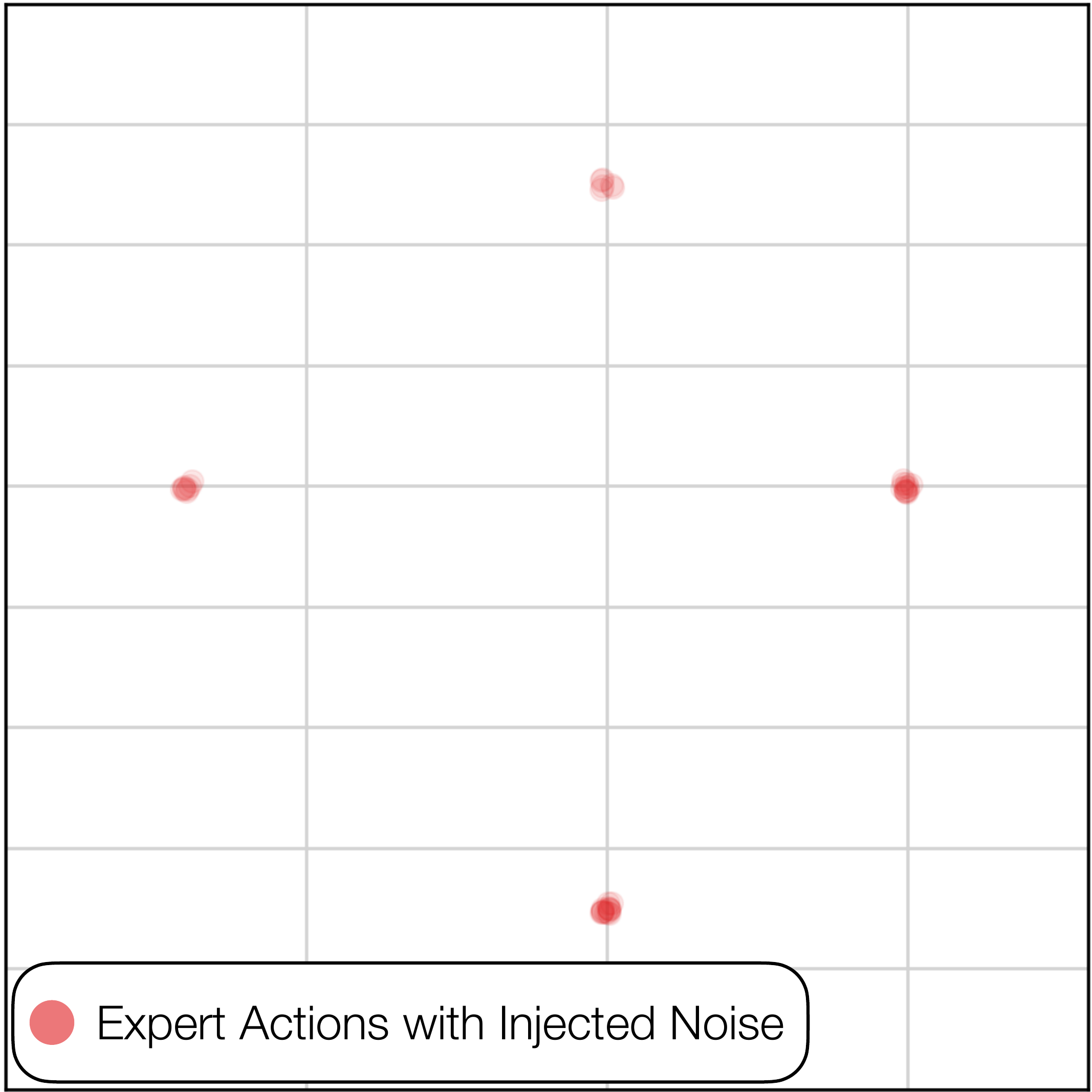}
        \caption{}
    \end{minipage}
    \hspace{0.15cm}
    \begin{minipage}[b]{0.23\textwidth}
        \centering
        \includegraphics[width=\textwidth, height=\textwidth]{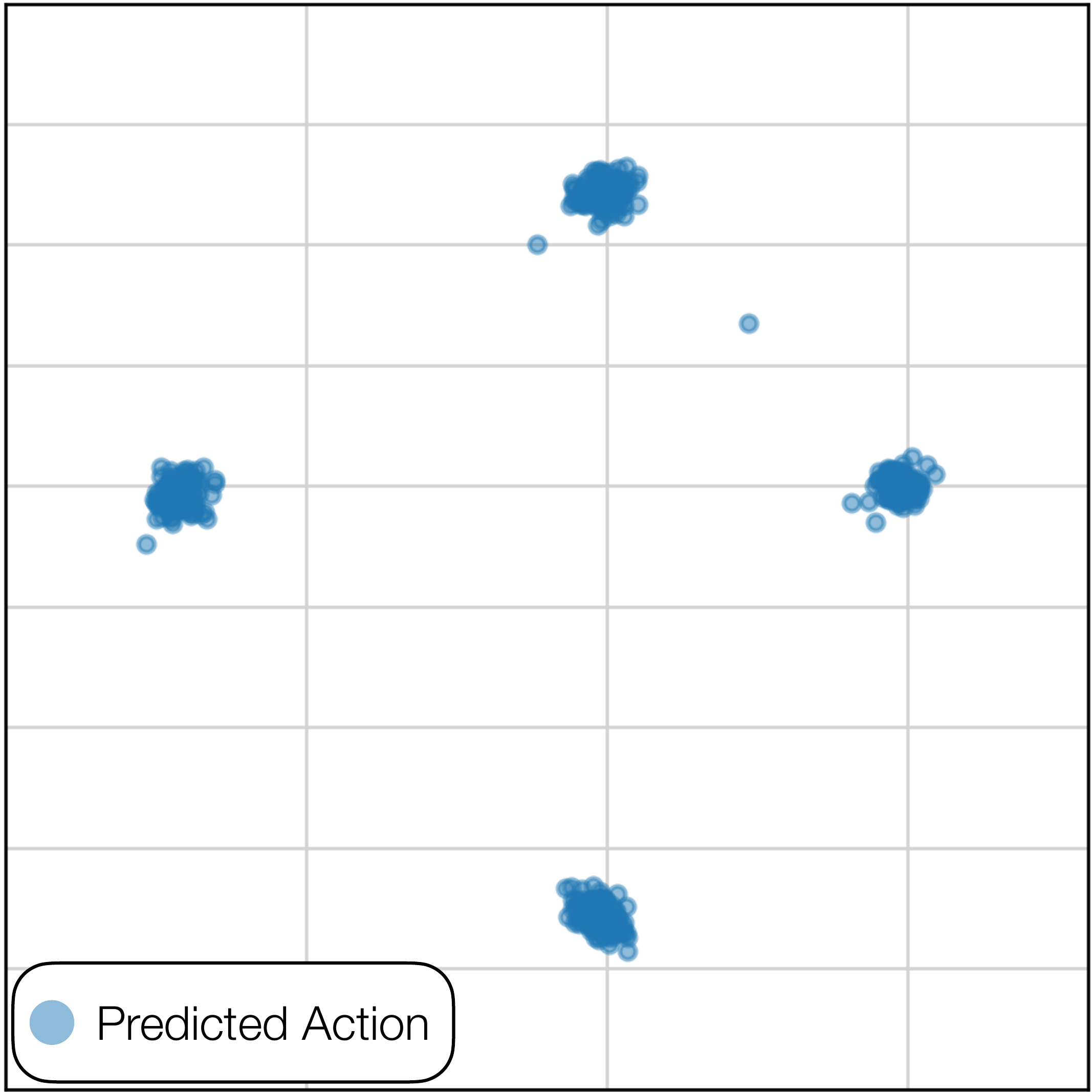}
        \caption{}
    \end{minipage}
    \hspace{0.15cm}
    \begin{minipage}[b]{0.23\textwidth}
        \centering
        \includegraphics[width=\textwidth, height=\textwidth]{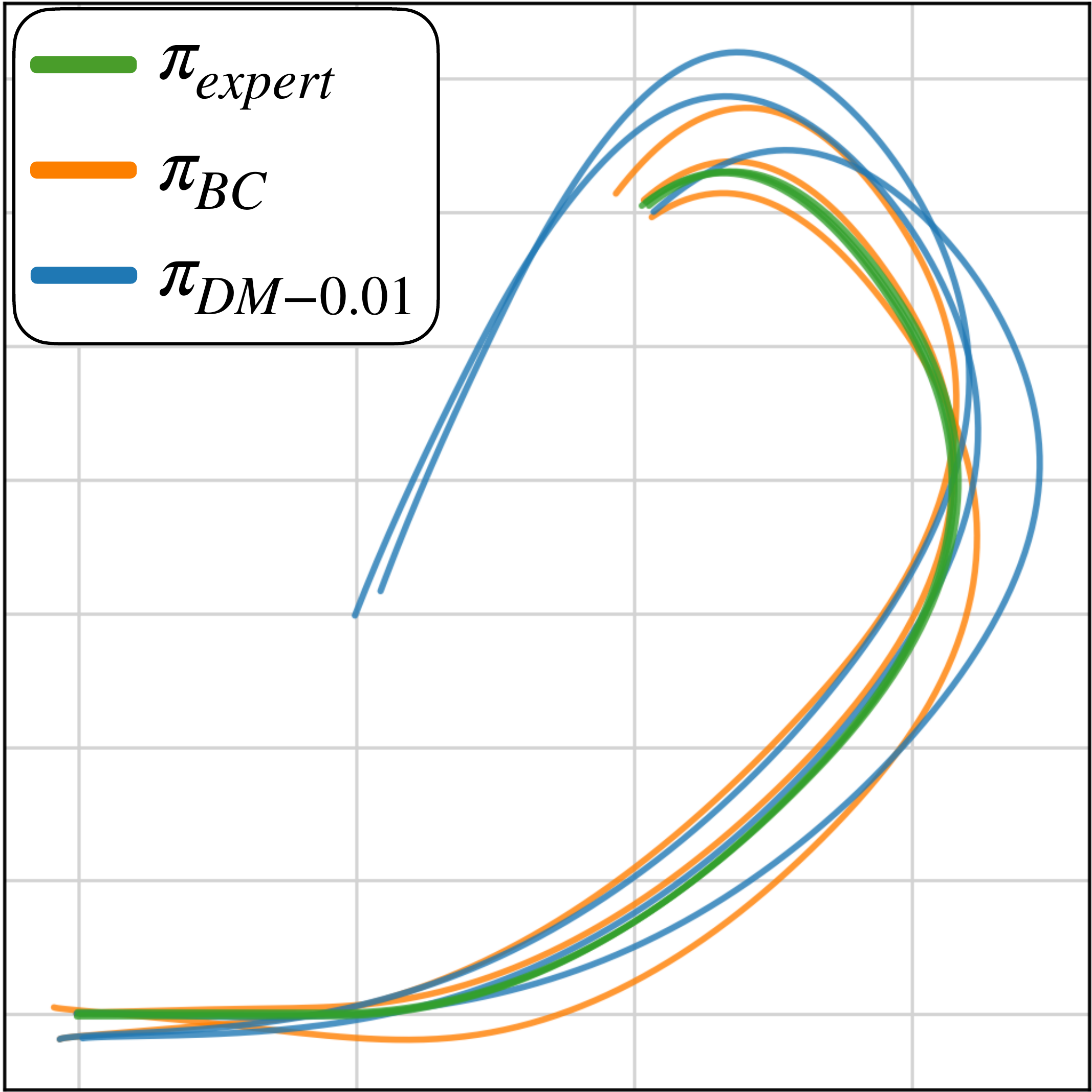}
        \caption{}
    \end{minipage}
    
    \caption[Comparing Modeling Conditional Probability and Joint Probability]{\textbf{Comparing Modeling Conditional Probability and Joint Probability.}  
    \textbf{(a) Expert actions.}
    \textbf{(b) Expert actions with injected noise.}
    \textbf{(c) Generated actions by the diffusion model.}
    \textbf{(d) Rollout trajectories.}
    }
    \label{fig:toy_noise}
\end{figure}

We evaluate the performance of the policies by rolling out each policy and calculating the distance between the end location of the policy and the expert end location. A policy rollout is considered successful if the distance is not greater than 0.1. 

In~\myfig{fig:toy_noise}, we visualize the expert actions, noise-injected expert actions, generated actions by the diffusion model trained with 0.01 noise level, and the rollout trajectories of the derived policy.
The result suggests that the diffusion model learning from expert distribution added with a preferable magnitude noise can better guide policy learning, achieving a success rate of 32\%, outperforming the original diffusion model that suffers more from the manifold overfitting with a success rate of 12\%. Yet, directly learning to model the conditional probability (i.e., $\pi_{\text{BC}}$) achieves a much higher success rate of 85\%.
This result verifies the advantage of modeling the conditional probability on this task, which motivates us to incorporate $\mathcal{L}_{\text{BC}}$ in our proposed learning objective instead of solely optimizing $\mathcal{L}_{\text{DM}}$.

\section{Effect of Dataset Size and Data Augmentation}
\label{sec:app_data}
\begin{table*}
\centering
\large
\caption[Varying Dataset Size in \maze{}]{\textbf{Varying Dataset Size in \maze{}.}
We report the mean and the standard deviation of the success rate of different dataset sizes of \maze{}. The results show that our proposed method DBC performs competitively against the Diffusion Policy and outperforms the other baselines across different dataset sizes. }

\scalebox{0.95}{\begin{tabular}{@{}ccccc@{}}\toprule
\multirow{2}{*}{\textbf{Method}} &
\multicolumn{4}{c}{\textbf{Dataset Size}} \\
& 25\% & 50\% & 75\% & 100\% \\
\cmidrule{1-5}
BC & 49.8\% $\pm$ 4.6\% 
& 71.9\% $\pm$ 4.9\% 
& 81.7\% $\pm$ 5.2\% 
& 92.1\% $\pm$ 3.6\% \\
Implicit BC 
& 51.9\% $\pm$ 3.7\%
& 65.9\% $\pm$ 5.1\% 
& 71.1\% $\pm$ 5.0\%
& 78.3\% $\pm$ 6.0\% \\
Diffusion Policy  
& \textbf{72.7}\% $\pm$ 9.2\%
& \textbf{83.7}\% $\pm$ 3.1\% 
& 88.4\% $\pm$ 4.5\% 
& \textbf{95.5}\% $\pm$ 1.9\% \\
\method{} (Ours)  
& \textbf{71.2}\% $\pm$ 3.9\%
& \textbf{83.9}\% $\pm$ 3.2\%
& \textbf{93.1}\% $\pm$ 2.6\% 
& \textbf{95.4}\% $\pm$ 1.7\% \\
\bottomrule
\end{tabular}}
\label{table:maze_dataset_size}
\end{table*}

In order to assess the impact of the training dataset's size, we conducted experiments in the \maze{} environment using 0.25, 0.5, and 0.75 fractions of the original dataset size.
The results in~\mytable{table:maze_dataset_size} show that our proposed method DBC performs competitively against the Diffusion Policy and outperforms the other baselines across different dataset sizes.
The BC baseline demonstrates a noticeable drop in performance as the dataset size decreases, and the Implicit BC baseline consistently exhibits inferior performance as the dataset size decreases.
The results demonstrate that our proposed framework and Diffusion Policy exhibit greater robustness to dataset size compared to BC and Implicit BC.

Since the size of the training dataset is important to the performance of the derived policy, we further evaluate if the learned diffusion model can be used for data augmentation.
We leverage the diffusion model learning from the expert dataset to generate state-action pairs as a data augmentation method.
Specifically, we use 18525 state-action pairs from the Maze dataset to train a diffusion model and then generate 18525 samples with the trained diffusion model. We combine the real and generated state-action pairs and then learn a BC policy.
The policy with data augmentation performs 2.06\% better than the one without data augmentation, where the improvement is within a standard deviation.
The above result is consistent with the dataset size experiment.

\section{Environment \& Task Details}
\label{sec:app_exp_task}
\subsection{\maze{}}
\label{sec:app_maze}

\myparagraph{Description} A point-maze agent in a 2D maze learns to navigate from its start location to a goal location by iteratively predicting its x and y acceleration. The 6D states include the agent's two-dimensional current location and velocity, and the goal location. The start and the goal locations are randomized when an episode is initialized.

\myparagraph{Evaluation} We evaluate the agents with $100$ episodes and three random seeds and compare our method with the baselines regarding the average success rate and episode lengths, representing the effectiveness and efficiency of the policy learned by different methods. An episode terminates when the maximum episode length of $400$ is reached. 

\myparagraph{Expert Dataset} The expert dataset consists of the $100$ demonstrations with $18,525$ transitions provided by~\citet{goalprox}.

\subsection{\fetchpick{}}
\label{sec:app_fetch}

\myparagraph{Description} \fetchpick{} requires a 7-DoF robot arm to pick up an object from the table and move it to a target location. Following the environment setups of~\citet{goalprox}, a 16D state representation consists of the angles of the robot joints, the robot arm poses relative to the object, and goal locations. The first three dimensions of the action indicate the desired relative position at the next time step. The fourth dimension of action specifies the distance between the two fingers of the gripper.

\myparagraph{Evaluation} We evaluate the agents with $100$ episodes and three random seeds and compare our method with the baselines regarding the average success rate and episode lengths. An episode terminates when the agent completes the task or the maximum episode length is reached, which is set to $50$ for \fetchpick{}.

\myparagraph{Expert Dataset} The expert dataset of \fetchpick{} consists of $303$ trajectories ($10k$ transitions) provided by~\citet{goalprox}.

\subsection{\handrotate{}}
\label{sec:app_hand}

\myparagraph{Description}
\handrotate{}~\citep{plappert2018multi} requires a 24-DoF Shadow Dexterous Hand to in-hand rotate a block to a target orientation. The 68D state representation consists of the joint angles and velocities of the hand, object poses, and the target rotation. The 20D action indicates the position control of the 20 joints, which can be controlled independently. \handrotate{} is extremely challenging due to its high dimensional state and action spaces. We adapt the experimental setup used in ~\citet{plappert2018multi} and ~\citet{goalprox}, where the rotation is restricted to the z-axis and the possible initial and target z rotations are set within $[-\frac{\pi}{12}, \frac{\pi}{12}]$ and $[\frac{\pi}{3}, \frac{2\pi}{3}]$, respectively. 

\myparagraph{Evaluation} 
We evaluate the agents with $100$ episodes and three random seeds and compare our method with the baselines regarding the average success rate and episode lengths. An episode terminates when the agent completes the goal or the maximum episode length of $50$ is reached.

\myparagraph{Expert Dataset} 
To collect expert demonstrations,
we train a SAC~\citep{haarnoja18b} policy using dense rewards for $10M$ environment steps. The dense reward given at each time step $t$ is $R(s_t, a_t) = d_t - d_{t+1}$, where $d_t$ and $d_{t+1}$ represent the angles (in radian) between current and the desired block orientations before and after taking the actions. 
Following the training stage, the SAC expert policy achieves a success rate of 59.48\%. Subsequently, we collect $515$ successful trajectories ($10k$ transitions) from this policy to form our expert dataset for \handrotate{}.

\subsection{\cheetah{}}
\label{sec:app_cheetah}

\myparagraph{Description} 
The \cheetah{} is a 2D robot with 17 states, indicating the status of each joint. The goal of this task is to exert torque on the joints to control the robot to walk toward x-direction. 
The agent would grant positive rewards for forward movement and negative rewards for backward movement.

\myparagraph{Evaluation} 
We evaluate each learned policy with $30$ episodes and three random seeds and compare our method with the baselines regarding the average returns of episodes. The return of an episode is accumulated from all the time steps of an episode. 

\myparagraph{Expert Dataset} 
The expert dataset consists of $5$ trajectories with $5k$ state-action pairs provided by~\citet{walkerdemo}.

\subsection{\walker{}}
\label{sec:app_walker}

\myparagraph{Description} 
\walker{} requires an agent to walk toward x-coordinate as fast as possible while maintaining its balance. The 17D state consists of angles of joints, angular velocities of joints, and velocities of the x and z-coordinate of the top. The 6D action specifies the torques to be applied on each joint of the walker avatar. 

\myparagraph{Evaluation} 
We evaluate each learned policy with $30$ episodes and three random seeds and compare our method with the baselines regarding the average returns of episodes. The return of an episode is accumulated from all the time steps of an episode. An episode terminates when the agent is unhealthy (\ie ill conditions predefined in the environment) or the maximum episode length ($1000$) is reached. 

\myparagraph{Expert Dataset} 
The expert dataset consists of $5$ trajectories with $5k$ state-action pairs provided by~\citet{walkerdemo}.

\subsection{\antreach{}}
\label{sec:app_ant}

\myparagraph{Description} 

In the \antreach{}, the task involves an ant robot with four legs aiming to reach a randomly generated goal on a half-circle with a 5-meter radius centered around the ant. The 42D state includes joint angles, velocities, and the relative position of the goal to the ant.
There is no noise added to the ant's initial pose during training. However, random noise is introduced during the evaluation, which requires the policies to generalize to unseen states.

\myparagraph{Evaluation} 
We evaluate the agents with $100$ episodes and three random seeds and compare our method with the baselines regarding the average success rate and episode lengths. An episode terminates when the agent completes the goal or the maximum episode length of $50$ is reached.

\myparagraph{Expert Dataset} 
The expert dataset comprises 10k state-action pairs provided by \citet{goalprox}.

\section{Model Architecture}
\label{sec:app_model}

\begin{table*}
\centering
\caption[Model Architectures]{\textbf{Model Architectures.}
We report the architectures used for all the methods on all the tasks.
}
\scalebox{0.85}{\begin{tabular}{@{}ccccccccc@{}}\toprule
\textbf{Method} & 
\textbf{Models} & 
\textbf{Component} &
\maze{} &
\fetchpick{} &
\handrotate{} &
\cheetah{} &
\walker{} &
\antreach{}
\\
\cmidrule{1-9}
\multirow{4}{*}{BC}
& \multirow{4}{*}{Policy $\pi$} 
& \# Layers & 4 & 3 & 3 & 3 & 3 & 3\\
& & Input Dim. & 6 & 16 & 68 & 17 & 17 & 42\\
& & Hidden Dim. & 256 & 750 & 512 & 256 & 1024 & 1024\\
& & Output Dim. & 2 & 4 & 20 & 6 & 6 & 8\\
\cmidrule{1-9}
\multirow{4}{*}{Implicit BC}
& \multirow{4}{*}{Policy $\pi$} 
& \# Layers & 2 & 2 & 2 & 3 & 2 & 2\\
& & Input Dim. & 8 & 20 & 88 & 23 & 23 & 50\\
& & Hidden Dim. & 1024 & 1024 & 512 & 512 & 1024 & 1200\\
& & Output Dim. & 1 & 1 & 1 & 1 & 1 & 1\\
\cmidrule{1-9}
\multirow{4}{*}{Diffusion Policy}
& \multirow{4}{*}{Policy $\pi$} 
& \# Layers & 5 & 5 & 5 & 5 & 5 & 5\\
& & Input Dim. & 8 & 20 & 88 & 23 & 23 & 42\\
& & Hidden Dim. & 256 & 1200 & 2100 & 1200 & 1200 & 1200\\
& & Output Dim. & 2 & 4 & 20 & 6 & 6 & 8\\
\cmidrule{1-9}
\multirow{8}{*}{\method{}} 
& \multirow{4}{*}{DM $\phi$} 
& \# Layers & 5 & 5 & 5 & 5 & 5 & 5\\
& & Input Dim. & 8 & 20 & 88 & 23 & 23 & 50\\
& & Hidden Dim. & 128 & 1024 & 2048 & 1024 & 1024 & 1024\\
& & Output Dim. & 8 & 20 & 88 & 23 & 23 & 50\\
\cmidrule{2-9}
& \multirow{4}{*}{Policy $\pi$} 
& \# Layers & 4 & 3 & 3 & 3 & 3 & 3\\
& & Input Dim. & 6 & 16 & 68 & 17 & 17 & 42\\
& & Hidden Dim. & 256 & 750 & 512 & 256 & 1024 & 1024\\
& & Output Dim. & 2 & 4 & 20 & 6 & 6 & 8\\
\bottomrule
\end{tabular}}
\label{table:architecture}
\end{table*}

This section describes the model architectures used for all the experiments. 
\mysecref{sec:app_model_main} presents the 
model architectures of BC, Implicit BC, Diffusion Policy, and our proposed framework \method{}.
\mysecref{sec:app_model_gm} details the 
model architectures of the EBM, VAE, and GAN 
used for the experiment comparing different generative models.

\subsection{Model Architecture of 
BC, Implicit BC, Diffusion Policy, and \method{}}
\label{sec:app_model_main}

We compare our \method{} with three baselines (BC, Implicit BC, and Diffusion Policy) on various tasks in~\mysecref{sec:exp_result}.
We detail the model architectures for all the methods on all the tasks in~\mytable{table:architecture}.
Note that all the models, the policy of BC, the energy-based model of Implicit BC, the conditional diffusion model of Diffusion Policy, the policy and the diffusion model of \method{}, are parameterized by a multilayer perceptron (MLP).
We report the implementation details for each method as follows.

\myparagraph{BC}
The non-linear activation function is a hyperbolic tangent for all the BC policies.
We experiment with BC policies with more parameters, which tend to severely overfit expert datasets, resulting in worse performance.

\myparagraph{Implicit BC}
The non-linear activation function is ReLU for all energy-based models of Implicit BC.
We empirically find that Implicit BC prefers shallow architectures in our tasks, so we set the number of layers to $2$ for the energy-based models.

\myparagraph{Diffusion Policy}
The non-linear activation function is ReLU for all the policies of Diffusion Policy. We empirically find that Diffusion Policy performs better with a deeper architecture. Therefore, we set the number of layers to $5$ for the policy. 
In most cases, we use a Diffusion Policy with more parameters than the total parameters of \method{} consisting of the policy and the diffusion model.

\myparagraph{\method{}}
The non-linear activation function is ReLU for the diffusion models and is a hyperbolic tangent for the policies.
We apply batch normalization and dropout layers with a $0.2$ ratio for the diffusion models on \fetchpick{}.

\subsection{Model Architecture of EBM, VAE, and GAN}
\label{sec:app_model_gm}
We compare different generative models (\ie EBM, VAE, and GAN) on \maze{} in~\mysecref{sec:app_generative_models}, and we report the model architectures used for the experiment in this section.

\myparagraph{Energy-Based Model}
An energy-based model (EBM) consists of $5$ linear layers with ReLU activation.
The EBM takes a concatenated state-action pair with a dimension of $8$ as input; the output is a $1$-dimensional vector representing the estimated energy values of the state-action pair.
The size of the hidden dimensions is $128$.

\myparagraph{Variational Autoencoder}
The architecture of a variational autoencoder consists of an encoder and a decoder.
The inputs of the encoder are a concatenated state-action pair, and the outputs are the predicted mean and variance,
which parameterize a Gaussian distribution. 
We apply the reparameterization trick~\citep{kingma2013auto}, sample features from the predicted Gaussian distribution, and use the decoder to produce the reconstructed state-action pair.
The encoder and the decoder both consist of $5$ linear layers with LeakyReLU~\citep{xu2020reluplex} activation.
The size of the hidden dimensions is $128$. 
That said, the encoder maps an $8$-dimensional state-action pair to two $128$-dimensional vectors (\ie mean and variance), and the decoder maps a sampled $128$-dimensional vector back to an $8$-dimensional reconstructed state-action pair.

\myparagraph{Generative Adversarial Network}
The architecture of the generative adversarial network consists of a generator and a discriminator.
The generator is the policy model that predicts an action from a given state, whose input dimension is $6$ and output dimension is $2$.
On the other hand, the discriminator learns to distinguish the expert state-action pairs $(s, a)$ from the state-action pairs produced by the generator $(s, \hat{a})$.
Therefore, the input dimension of the discriminator is $8$, and the output is a scalar representing the probability of the state-action pair being "real."
The generator and the discriminator both consist of three linear layers with ReLU activation, and the size of the hidden dimensions is $256$. 

\section{Training and Inference Details}
\label{sec:app_training_evaluation}

We describe the details of training and performing inference in this section, 
including computation resources and hyperparameters.

\subsection{Computation Resource}
\label{sec:app_copute}

We conducted all the experiments on the following three workstations:
\begin{itemize}
    \item M1: ASUS WS880T workstation with an Intel Xeon W-2255 (10C/20T, 19.25M, 4.5GHz) 48-Lane CPU, 64GB memory, an NVIDIA RTX 3080 Ti GPU, and an NVIDIA RTX 3090 Ti GPU
    \item M2: ASUS WS880T workstation with an Intel Xeon W-2255 (10C/20T, 19.25M, 4.5GHz) 48-Lane CPU, 64GB memory, an NVIDIA RTX 3080 Ti GPU, and an NVIDIA RTX 3090 Ti GPU   
    \item M3: ASUS WS880T workstation with an Intel Xeon W-2255 (10C/20T, 19.25M, 4.5GHz) 48-Lane CPU, 64GB memory, and two NVIDIA RTX 3080 Ti GPUs    
\end{itemize}

\subsection{Hyperparamters}
\label{sec:app_hyperparamters}

We report the hyperparameters used for all the methods on all the tasks in~\mytable{table:hyperparameter}.
We use the Adam optimizer~\citep{kingma2014adam} for all the methods on all the tasks and use linear learning rate decay for all policy models.

\begin{table*}
\centering
\small
\caption[Hyperparameters]{\textbf{Hyperparameters.}
This table reports the hyperparameters used for all the methods on all the tasks. Note that our proposed framework (\method{}) consists of two learning modules, the diffusion model and the policy, and therefore their hyperparameters are reported separately.}
\scalebox{0.95}{\begin{tabular}{@{}cccccccc@{}}\toprule
\textbf{Method} & \textbf{Hyperparameter} & 
\maze{} &
\fetchpick{} &
\handrotate{} &
\cheetah{} &
\walker{} &
\antreach{} \\
\cmidrule{1-8}
\multirow{3}{*}{BC} & Learning Rate & 5e-5 & 5e-6 & 1e-4 & 1e-4 & 1e-4 & 0.006\\
& Batch Size & 128 & 128 & 128 & 128 & 128 & 128 \\
& \# Epochs & 2000 & 5000 & 5000 & 1000 & 1000 & 10000\\
\cmidrule{1-8}
\multirow{3}{*}{Implicit BC} & Learning Rate & 1e-4 & 5e-6 & 1e-5 & 1e-4 & 8e-5 & 5e-5\\
& Batch Size & 128 & 512 & 128 & 128 & 128 & 128 \\
& \# Epochs & 10000 & 15000 & 15000 & 10000 & 10000 & 30000\\
\cmidrule{1-8}
\multirow{3}{*}{Diffusion Policy} & Learning Rate & 2e-4 & 1e-5 & 1e-4 & 1e-4 & 1e-4 & 1e-5\\
& Batch Size & 128 & 128 & 128 & 128 & 128 & 128\\
& \# Epochs & 20000 & 15000 & 30000 & 10000 & 10000 & 30000 \\
\cmidrule{1-8}
\multirow{7}{*}{\method{} (Ours)}
& Diffusion Model Learning rate & 1e-4 & 1e-3 & 3e-5 & 2e-4 & 2e-4 & 2e-4\\
& Diffusion Model Batch Size & 128 & 128 & 128 & 128 & 128 & 1024\\
& Diffusion Model \# Epochs & 8000 & 10000 & 10000 & 8000 & 8000 & 20000 \\
\cmidrule{2-8}
& Policy Learning Rate & 5e-5 & 5e-6 & 1e-4 & 1e-4 & 1e-4 & 0.006\\
& Policy Batch Size & 128 & 128 & 128 & 128 & 128 & 128 \\
& Policy \# Epochs & 2000 & 5000 & 5000 & 1000 & 1000 & 10000 \\
& $\lambda$ & 30 & 0.5 & 10 & 0.2 & 0.2 & 1\\
\bottomrule
\end{tabular}}
\label{table:hyperparameter}
\end{table*}

\subsection{Inference Details}
\label{sec:app_evaluation}

This section describes how each method infers an action $\hat{a}$ given a state $s$.

\myparagraph{BC \& \method{}}
The policy models of BC and \method{} can directly predict an action given a state, \ie $\hat{a} \sim \pi(s)$, and are therefore more efficient during inference as described in~\mysecref{sec:exp_result}.

\myparagraph{Implicit BC}
The energy-based model (EBM) of Implicit BC learns to predict an estimated energy value for a state-action pair during training.
To generate a predicted $\hat{a}$ given a state $s$ during inference, it requires a procedure to sample and optimize actions.
We follow \citet{florence2022implicit} and implement a derivative-free optimization algorithm to perform inference.

The algorithm first randomly samples $N_{s}$ vectors from the action space as candidates. 
The EBM then produces the estimated energy value of each candidate action and applies the Softmax function on the estimated energy values to produce a $N_{s}$-dimensional probability.
Then, it samples candidate actions according to the above probability and adds noise to them to generate another $N_{s}$ candidates for the next iteration.
The above procedure iterates $N_{iter}$ times.
Finally, the action with maximum probability in the last iteration is selected as the predicted action $\hat{a}$.
In our experiments, $N_{s}$ is set to $1000$ and $N_{iter}$ is set to $3$.

\myparagraph{Diffusion Policy}
Diffusion Policy learns a conditional diffusion model as a policy and produces an action from sampled noise vectors conditioning on the given state during inference.
We follow~\citet{pearce2022imitating, chi2023diffusionpolicy} and adopt Denoising Diffusion Probabilistic Models (DDPMs)~\citep{ho2020ddpm} for the diffusion models.
Once learned, the diffusion policy $\pi$ can 
"denoise" a noise sampled from a Gaussian distribution $\mathcal{N}(0, 1)$ given a state $s$ and yield a predicted action $\hat{a}$ using the following equation: 
\begin{equation}
\label{eq:rev_diff} 
a_{n-1} = \frac{1}{\sqrt{\alpha_{n}}}(a_n-\frac{1-\alpha_{n}}{\sqrt{1-\Bar{\alpha}_n}}\pi(s, a_n, n)) + \sigma_nz,
\end{equation}
where $\alpha_n$, $\Bar{\alpha}_n$, and $\sigma_n$ are schedule parameters, $n$ is the current time step of the reverse diffusion process, and $z \sim \mathcal{N}(0, 1)$ is a random vector.
The above denoising process iterates $N$ times to produce a predicted action $a_0$ from a sampled noise $a_N \sim \mathcal{N}(0, 1)$.
The number of total diffusion steps $N$ is $1000$ in our experiment, which is the same for the diffusion model in \method{}.

\subsection{Training Details of Generative Models}
\label{sec:app_gm}
Our proposed framework employs a diffusion model (DM) to model the joint probability of expert state-action pairs and utilizes it to guide policy learning.
We justify our choice of generative models by using popular models, including energy-based models (EBMs)~\citep{du2019implicit, song2021train}, variational autoencoders (VAEs)~\citep{kingma2013auto}, and generative adversarial networks (GANs)~\citep{goodfellow2014generative} as well as the diffusion model in~\mysecref{sec:app_generative_models}.
Each generative model learns to model the joint distribution of expert state-action pairs.
For fair comparisons, all the policy models learning from learned generative models consists of $3$ linear layers with ReLU activation, where the hidden dimension is $256$.
All the policies are trained for $2000$ epochs using the Adam optimizer~\citep{kingma2014adam}, and a linear learning rate decay is applied for EBMs and VAEs.
The following sections detail the training of generative models and the approaches to guide policy learning.

\subsubsection{Energy-Based Model}
\label{sec:app_gm_ebm}
\myparagraph{Model Learning}
Energy-based models (EBMs) learn to model the joint distribution of the expert state-action pairs by predicting an estimated energy value for a state-action pair $(s, a)$. The EBM aims to assign low energy value to the real expert state-action pairs while high energy otherwise. Therefore, the predicted energy value can be used to evaluate how well a state-action pair $(s, a)$ fits the distribution of the expert state-action pair distribution.

To train the EBM, we generate $N_{neg}$ random actions as negative samples for each expert state-action pair as proposed in~\citet{florence2022implicit}.
The objective of the EBM $E_{\phi}$ is the InfoNCE loss \citep{oord2018representation}:
\begin{equation}
\label{eq:emb_loss} 
\mathcal{L}_{\text{InfoNCE}} = \frac{e^{-E_{\phi}(s, a)}}{e^{-E_{\phi}(s, a)}+\Sigma_{i=1}^{N_{neg}}e^{-E_{\phi}(s, \tilde{a}_{i})}},
\end{equation}
where $(s, a)$ indicates an expert state-action pair, $\tilde{a}_i$ indicates the sampled random action, and $N_{neg}$ is set to $64$ in our experiments.
The EBM learns to separate the expert state-action pairs from the negative samples by optimizing the above InfoNCE loss. 

The EBM is trained for $8000$ epochs with the Adam optimizer~\citep{kingma2014adam}, with a batch size of $128$ and an initial learning rate of $0.0005$.
We apply learning rate decay by $0.99$ for every $100$ epoch.

\myparagraph{Guiding Policy Learning}
To guide a policy $\pi$ to learn, we design an EBM loss
$\mathcal{L}_{\text{EBM}} = E_{\phi}(s, \hat{a})$, where $\hat{a}$ indicates the predicted action produced by the policy.
The above EBM loss regularizes the policy to generate actions with low energy values, which encourage the predicted state-action pair $(s, \hat{a})$ to fit the modeled expert state-action pair distribution.
The policy learning from this EBM loss $\mathcal{L}_{\text{EBM}}$ achieves a success rate of $49.09\%$ in \maze{} as reported in~\mytable{table:gm}.

We also experiment with combining this 
EBM loss $\mathcal{L}_{\text{EBM}}$ with the 
$\mathcal{L}_{\text{BC}}$ loss.
The policy optimizes $\mathcal{L}_{\text{BC}}+
\lambda_{\text{EBM}}\mathcal{L}_{\text{EBM}}$, where $\lambda_{\text{EBM}}$ is set to $0.1$. 
Optimizing this combined loss yields a success rate of $80.00\%$ in \maze{} as reported in~\mytable{table:gm}.

\subsubsection{Variational Autoencoder}
\label{sec:app_gm_vae}

\myparagraph{Model Learning}
Variational autoencoders (VAEs) model the joint distribution of the expert data by learning to reconstruct expert state-action pairs $(s, a)$.
Once the VAE is learned, how well a state-action pair fits the expert distribution can be reflected in the reconstruction loss.

The objective of training a VAE is as follows:
\begin{equation}
\label{eq:vae_loss} 
\mathcal{L}_{\text{vae}} = {||\hat{x} - x||}^2 + D_{\text{KL}}(\mathcal{N}(\mu_{x}, \sigma_{x})||\mathcal{N}(0, 1)),
\end{equation}
where $x$ is the latent variable, \ie the concatenated state-action pair $x = [s, a]$, and $\hat{x}$ is the reconstruction of $x$, \ie the reconstructed state-action pair.
The first term is the reconstruction loss, 
while the second term encourages aligning the data distribution with a normal distribution $\mathcal{N}(0, 1)$, where $\mu_{x}$ and $\sigma_{x}$ are the predicted mean and standard deviation given $x$.

The VAE is trained for $100k$ update iterations with the Adam optimizer~\citep{kingma2014adam},
with a batch size of $128$ and an initial learning rate of $0.0001$.
We apply learning rate decay by $0.5$ for every $5k$ epoch.

\myparagraph{Guiding Policy Learning}
To guide a policy $\pi$ to learn, we design a VAE loss
$\mathcal{L}_{\text{VAE}} = max(\mathcal{L}_{\text{vae}}^{\text{agent}}-\mathcal{L}_{\text{vae}}^{\text{expert}}, 0)$, similar to~\myeq{eq:dm_loss}.
This loss forces the policy to predict an action, together with the state, that can be well reconstructed with the learned VAE.
The policy learning from this VAE loss $\mathcal{L}_{\text{VAE}}$ achieves a success rate of $48.47\%$ in \maze{} as reported in~\mytable{table:gm}.

We also experiment with combining this 
VAE loss $\mathcal{L}_{\text{VAE}}$ with the 
$\mathcal{L}_{\text{BC}}$ loss.
The policy optimizes $\mathcal{L}_{\text{BC}}+
\lambda_{\text{VAE}}\mathcal{L}_{\text{VAE}}$, where $\lambda_{\text{VAE}}$ is set to $1$. 
Optimizing this combined loss yields a success rate of $82.31\%$ in \maze{} as reported in~\mytable{table:gm}.

\subsubsection{Generative Adversarial Network}
\label{sec:app_gm_gan}
\myparagraph{Adversarial Model Learning \& Policy Learning}
Generative adversarial networks (GANs) model the joint distribution of expert data with a generator and a discriminator.
The generator aims to synthesize a predicted action $\hat{a}$ given a state $s$.
On the other hand, the discriminator aims to identify expert the state-action pair $(s, a)$ from the predicted one $(s, \hat{a})$.
Therefore, a learned discriminator can evaluate how well a state-action pair fits the expert distribution.

While it is possible to learn a GAN separately and utilize the discriminator to guide policy learning, we let the policy $\pi$ be the generator directly and optimize the policy with the discriminator iteratively.
We hypothesize that a learned discriminator may be too selective for policy training from scratch, so we learn the policy $\pi$ with the discriminator $D$ to improve the policy and the discriminator simultaneously.

The objective of training the discriminator $D$ is as follows:
\begin{equation}
\label{eq:disc_loss} 
\mathcal{L}_{\text{disc}} = BCE(D(s, a), 1) + BCE(D(s, \hat{a}), 0) = -log(D(s, a)) - log(1-D(s, \hat{a})),
\end{equation}
where $\hat{a} = \pi(s)$ is the predicted action, and $BCE$ is the binary cross entropy loss.
The binary label $(0, 1)$ indicates whether or not the state-action pair sampled from the expert data.
The generator and the discriminator are both updated by Adam optimizers using a $0.00005$ learning rate.

To learn a policy (\ie generator), we design the following GAN loss:
\begin{equation}
\label{eq:gan_loss} 
\mathcal{L}_{\text{GAN}} = BCE(D(s, \hat{a}), 1) = -log(D(s, \hat{a})).
\end{equation}
The above GAN loss guides the policy to generate state-action pairs that fit the joint distribution of the expert data.
The policy learning from this GAN loss $\mathcal{L}_{\text{GAN}}$ achieves a success rate of $50.29\%$ in \maze{} as reported in~\mytable{table:gm}.

We also experiment with combining this 
GAN loss $\mathcal{L}_{\text{GAN}}$ with the 
$\mathcal{L}_{\text{BC}}$ loss.
The policy optimizes $\mathcal{L}_{\text{BC}}+
\lambda_{\text{GAN}}\mathcal{L}_{\text{GAN}}$, where $\lambda_{\text{GAN}}$ is set to $0.2$. 
Optimizing this combined loss yields a success rate of $71.64\%$ in \maze{} as reported in~\mytable{table:gm}.

\section{On the Theoretical Motivation for Guiding Policy Learning with Diffusion Model}
\label{sec:app_diffusion_model}

This section further elaborates on the technical motivation for leveraging diffusion models for imitation learning. Specifically, we aim to learn a diffusion model to model the joint distribution of expert state-action pairs. 
Then, we propose to utilize this learned diffusion model to augment a BC policy that aims to imitate expert behaviors. 

We consider the distribution of expert state-action pairs as the real data distribution $q_x$ in learning a diffusion model. 
Following this setup, $x_0$ represents an original expert state-action pair $(s, a)$ and $q(x_n|x_{n-1})$ represents the forward diffusion process, which gradually adds Gaussian noise to the data in each timestep $n = 1, ..., N$ until $x_N$ becomes an isotropic gaussian distribution. 
On the other hand, the reverse diffusion process is defined as $\phi(x_{n-1}|x_n) := \mathcal{N}(x_{n-1}; \mu_{\theta}(x_n, n), \Sigma_{\theta}(x_n, n))$, where $\theta$ denotes the learnable parameters of the diffusion model $\phi$, as illustrated in~\myfig{fig:dm}.

Our key idea is to use the proposed diffusion model loss $\mathcal{L}_{\text{DM}}$ in~\myeq{eq:dm_loss} as an estimate of how well a predicted state-action pair $(s,\hat{a})$ fits the expert state-action pair distribution, as described in~\mysecref{sec:dm_loss}. 
In the following derivation, we will show that by optimizing this diffusion model loss $\mathcal{L}_{\text{DM}}$, we maximize the lower bound of the agent data’s probability under the derived expert distribution and hence bring the agent policy $\pi$ closer to the expert policy $\pi^E$, which is the goal of imitation learning.

As depicted in ~\citet{luo2022understanding}, one can conceptualize diffusion models, including DDPM ~\citep{ho2020ddpm} adopted in this work,  as a hierarchical variational autoencoder~\citep{kingma2013auto}, which maximizes the likelihood $p(x)$ of observed data points $x$. 
Therefore, similar to hierarchical variational autoencoders, diffusion models can optimize the Evidence Lower Bound (ELBO) by minimizing the KL divergence $D_{KL}(q(x_{n-1}|x_n,x_0)||\phi(x_{n-1}|x_n))$.
Consequently, this can be viewed as minimizing the KL divergence to fit the distribution of the predicted state-action pairs $(s,\hat{a})$ to the distribution of expert state-action pairs.

According to Bayes’ theorem and the properties of Markov chains, the forward diffusion process $q(x_{n-1}|x_n,x_0)$ follows: 
\begin{equation}
\begin{aligned}
   q(x_{n-1}|x_n,x_0) \sim \mathcal{N}(x_{n-1}; 
   & \underbrace{\frac {\sqrt{\alpha_n} (1-\bar{\alpha}_{n-1})x_n + \sqrt{\bar{\alpha}_{n-1}} (1-\alpha_n) x_0} {1-\bar{\alpha_n}}}{\mu_q(x_n,x_0)}, \\
   & \underbrace{\frac {(1-\alpha_n) (1-\bar{\alpha}_{n-1})} {1-\bar{\alpha}_n}}{\Sigma_q(n)}).
\end{aligned}
\end{equation}
The variation term $\Sigma_q(n)$ in the above equation can be written as $\sigma^2_q(n)I$, where $\sigma^2_q(n)=\displaystyle \frac{(1-\alpha_n)(1-\bar{\alpha}_{n-1})}{1-\bar{\alpha}_n}$.
Therefore, minimizing the KL divergence is equivalent to minimizing the gap between the mean values of the two distributions:
\begin{equation}
\begin{aligned}
& \mathop{\arg\min}_{\theta} D_{KL}(q(x_{n-1}|x_n,x_0)||\phi(x_{n-1}|x_n)) \\
&=\mathop{\arg\min}_{\theta} D_{KL}(\mathcal{N}(x_{n-1};\mu_q,\Sigma_q(n))||\mathcal{N}(x_{n-1};\mu_\theta,\Sigma_q(n))) \\
&=\mathop{\arg\min}_{\theta} \frac {1}{2\sigma^2_q(n)}[||\mu_\theta-\mu_q||^2_2], 
\end{aligned}
\end{equation}
where $\mu_q$ represents the denoising transition mean and $\mu_\theta$ represents the approximated denoising transition mean by the model. 

Different implementations adopt different forms to model $\mu_\theta$. Specifically, for DDPMs adopted in this work, the true denoising transition means $\mu_q(x_n, x_0)$ derived above can be rewritten as: 
\begin{equation}
\begin{aligned}
\mu_q(x_n, x_0) = \frac{1}{\sqrt{\alpha_n}} (x_n - \frac{1-\alpha_n}{\sqrt{1-\bar{\alpha}_n}}\epsilon_0),
\end{aligned}
\end{equation}
which is referenced from Eq. 11 in~\citet{ho2020ddpm}. 
Hence, we can set our approximate denoising transition mean $\mu_\theta$ in the same form as the true denoising transition mean:
\begin{equation}
\begin{aligned}
\label{eq:dsde}
\mu_\theta(x_n, n) = \frac{1}{\sqrt{\alpha_n}}(x_n - \frac {1-\alpha_n}{\sqrt{1-\bar{\alpha}_n}}\hat{\epsilon_\theta}(x_n,n)), 
\end{aligned}
\end{equation}
as illustrated in~\citet{popov2021diffusion}. ~\citet{song2020score} further shows that the entire diffusion model formulation can be revised to view continuous stochastic differential equations (SDEs) as a forward diffusion. It points out that the reverse process is also an SDE, which can be computed by estimating a score function $\nabla_x \log p_t(x)$ at each denoising time step.
The idea of representing a distribution by modeling its score function is introduced in~\citet{song2019generative}. The fundamental concept is to model the gradient of the log probability density function $\nabla_x \log p_t(x)$, a quantity commonly referred to as the (Stein) score function. Such score-based models are not required to have a tractable normalizing constant and can be directly acquired through score matching. 
The measure of this score function determines the optimal path to take in the space of the data distribution to maximize the log probability under the derived real distribution. 

% \begin{figure}[t]
%     \centering
%     \begin{subfigure}[b]{0.3\textwidth}
%     \centering
%     \includegraphics[width=\textwidth, height=\textwidth]{figure/maze_env.png}
%     \caption{\textbf{Maze Layout}}
%     \label{fig:maze_env}
%     \end{subfigure}
%     \hspace{2 cm}
%     \begin{subfigure}[b]{0.3\textwidth}
%     \centering
%     \includegraphics[width=\textwidth, height=\textwidth]{figure/gradient_field.png}
%     \caption{\textbf{Learned Gradient Field}}
%     \label{fig:gradient_field}
%     \end{subfigure}
%     \caption[Visualization of the Gradient Field.]
%     {\textbf{Visualization of the Gradient Field.} 
%     \textbf{(a) Maze Layout}:
%     The layout of the medium maze used for \maze{}.    
%     \textbf{(b) Learned Gradient Field}:
%     We visualize the \maze{} expert demonstration as a distribution of points by their first two dimensions in gray. The points that cluster densely have a high probability, and vice versa.
%     Once a diffusion model is well-trained, it can move randomly sampled points to the area with high probability by predicting gradients (blue arrows). 
%     Accordingly, the estimate $p(s,a)$ of joint distribution modeling can serve as guidance for policy learning, as proposed in this work.
%     }
%     \label{fig:SDE_gradient}
% \end{figure}

\begin{figure}[t]
    \centering
    
    \begin{minipage}[b]{0.35\textwidth}
        \centering
        \includegraphics[width=\textwidth, height=\textwidth]{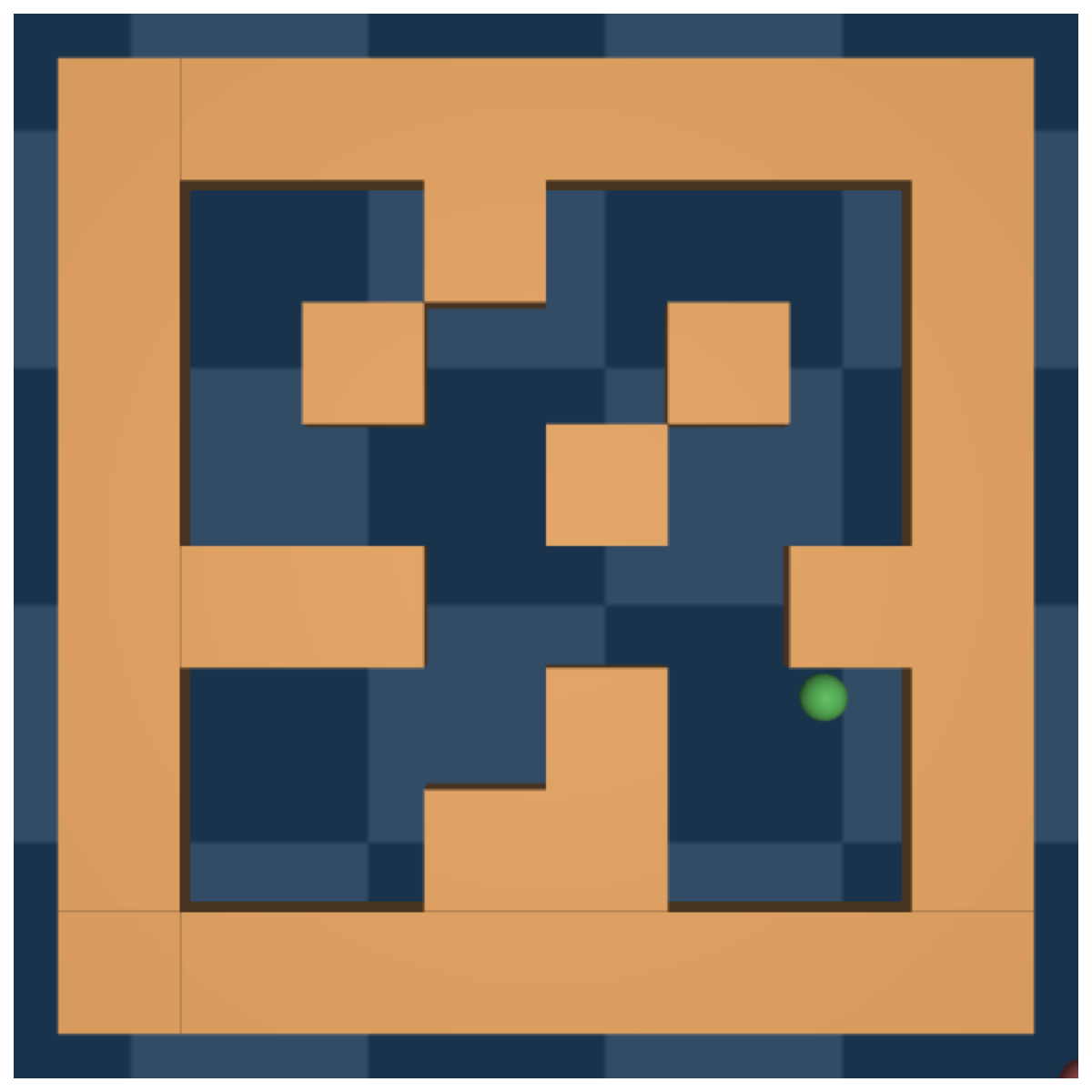}
        \subcaption{\textbf{Maze Layout}}
        \label{fig:maze_env}
    \end{minipage}
    \hspace{2cm}
    \begin{minipage}[b]{0.35\textwidth}
        \centering
        \includegraphics[width=\textwidth, height=\textwidth]{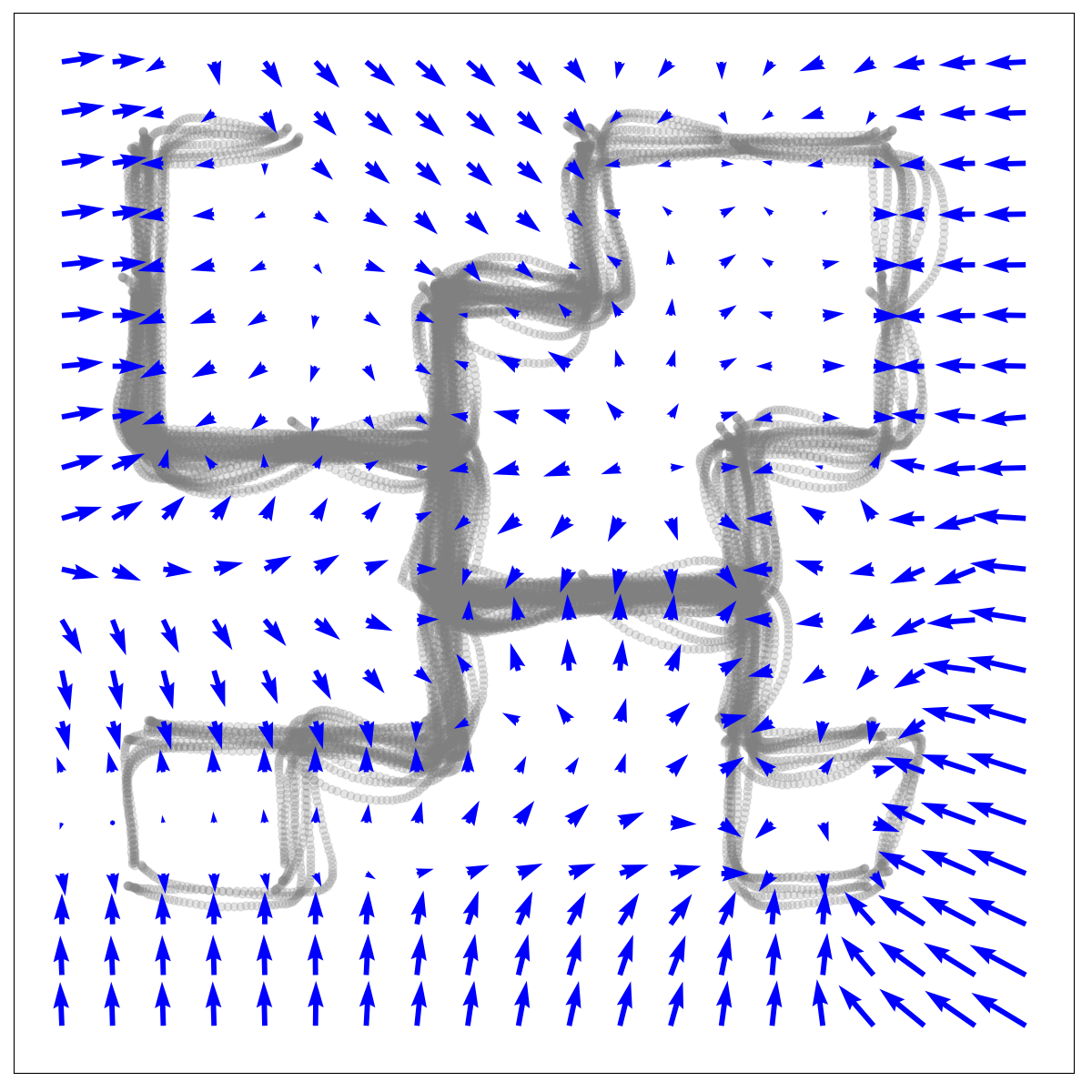}
        \subcaption{\textbf{Learned Gradient Field}}
        \label{fig:gradient_field}
    \end{minipage}
    
    \caption[Visualization of the Gradient Field.]
    {\textbf{Visualization of the Gradient Field.} 
    \textbf{(a) Maze Layout}:
    The layout of the medium maze used for \maze{}.    
    \textbf{(b) Learned Gradient Field}:
    We visualize the \maze{} expert demonstration as a distribution of points by their first two dimensions in gray. The points that cluster densely have a high probability, and vice versa.
    Once a diffusion model is well-trained, it can move randomly sampled points to the area with high probability by predicting gradients (blue arrows). 
    Accordingly, the estimate $p(s,a)$ of joint distribution modeling can serve as guidance for policy learning, as proposed in this work.
    }
    \label{fig:SDE_gradient}
\end{figure}

As shown in~\myfig{fig:gradient_field}, we visualized the learned gradient field of a diffusion model, which learns to model the expert state-action pairs in \maze{}.
Once trained, this diffusion model can guide a policy 
with predicted gradients (blue arrows) to move to areas with high probability, as proposed in our work.

Essentially, by moving in the opposite direction of the source noise, which is added to a data point $x_t$ to corrupt it, the data point is “denoised”; hence the log probability is maximized. This is supported by the fact that modeling the score function is the same as modeling the negative of the source noise. 
This perspective of the diffusion model is dubbed diffusion SDE. Moreover, \citet{popov2021diffusion} prove that~\myeq{eq:dsde} is diffusion SDE’s maximum likelihood SDE solver. Hence, the corresponding divergence optimization problem can be rewritten as:
\begin{equation}
\begin{aligned}
& \mathop{\arg\min}_{\theta} D_{KL}(q(x_{n-1}|x_n,x_0)||\phi(x_{n-1}|x_n)) \\
&=\mathop{\arg\min}_{\theta} \frac {1}{2\sigma^2_q(n)} \frac {(1-\alpha_n)^2}{(1-\bar{\alpha}_n)\alpha_n}[||\hat{\epsilon}_\theta(x_n,n) - \epsilon_0||^2_2],
\end{aligned}
\end{equation}
where $\epsilon_\theta$ is a function approximator aim to predict $\epsilon$ from $x$. As the coefficients can be omitted during optimization, we yield the learning objective $\mathcal{L}_{\text{diff}}$ as stated in in~\myeq{eq:diff_loss}:
\begin{equation}
\begin{aligned}
\mathcal{L}_{\text{diff}}(s, a, \phi) = \mathbb{E}_{n \sim N, (s, a) \sim D} \{{||\hat{\epsilon}(s, a, n) - \epsilon(n)||}^2 \} \\
= \mathbb{E}_{n \sim N, (s, a) \sim D}\{{||\phi(s, a, \epsilon(n)) - \epsilon(n)||}^2\}.
\end{aligned}
\end{equation}

The above derivation motivates our proposed framework that augments a BC policy by using the diffusion model to provide guidance that captures the joint probability of expert state-action pairs. Based on the above derivation, minimizing the proposed diffusion model loss (\ie learning to denoise) is equivalent to finding the optimal path to take in the data space to maximize the log probability. To be more accurate, when the learner policy predicts an action that obtains a lower $\mathcal{L}_{\text{diff}}$, it means that the predicted action $\hat{a}$, together with the given state $s$, fits better with the expert distribution. 

Accordingly, by minimizing our proposed diffusion loss, the policy is encouraged to imitate the expert policy.
To further alleviate the impact of rarely-seen state-action pairs $(s, a)$, we propose to compute the above diffusion loss for both expert data $(s, a)$ and predicted data $(s, \hat{a})$ and yield $\mathcal{L}_{\text{diff}}^{\text{expert}}$ and $\mathcal{L}_{\text{diff}}^{\text{agent}}$, respectively.
Therefore, we propose to augment BC with this objective: $ \mathcal{L}_{\text{DM}} = 
\mathbb{E}_{(s, a) \sim D, \hat{a} \sim \pi(s)}
\{max(\mathcal{L}_{\text{diff}}^{\text{agent}}-\mathcal{L}_{\text{diff}}^{\text{expert}}, 0)\}.$
%%%%%%%%%%%%%%%%%%%%%%%%%%%%%%%%%%%%%%%%%%%%%%%%%%%%%%%%%%%%%%%%%%%%%%%%%%%%%%%
%%%%%%%%%%%%%%%%%%%%%%%%%%%%%%%%%%%%%%%%%%%%%%%%%%%%%%%%%%%%%%%%%%%%%%%%%%%%%%%
%%%%%%%%%%%%%%%%%%%%%%%%%%%%%%%%%%%%%%%%%%%%%%%%%%%%%%%%%%%%%%%%%%%%%%%%%%%%%%%
%%%%%%%%%%%%%%%%%%%%%%%%%%%%%%%%%%%%%%%%%%%%%%%%%%%%%%%%%%%%%%%%%%%%%%%%%%%%%%%

% This document was modified from the file originally made available by
% Pat Langley and Andrea Danyluk for ICML-2K. This version was created
% by Iain Murray in 2018, and modified by Alexandre Bouchard in
% 2019 and 2021 and by Csaba Szepesvari, Gang Niu and Sivan Sabato in 2022.
% Modified again in 2023 and 2024 by Sivan Sabato and Jonathan Scarlett.
% Previous contributors include Dan Roy, Lise Getoor and Tobias
% Scheffer, which was slightly modified from the 2010 version by
% Thorsten Joachims & Johannes Fuernkranz, slightly modified from the
% 2009 version by Kiri Wagstaff and Sam Roweis's 2008 version, which is
% slightly modified from Prasad Tadepalli's 2007 version which is a
% lightly changed version of the previous year's version by Andrew
% Moore, which was in turn edited from those of Kristian Kersting and
% Codrina Lauth. Alex Smola contributed to the algorithmic style files.

\end{document}